%% file: main.tex
\documentclass[lettersize,journal]{IEEEtran}
\usepackage{amsmath,amsfonts}

\usepackage{array}
\usepackage{textcomp}
\usepackage{stfloats}
\usepackage{url}
\usepackage{verbatim}
\usepackage{graphicx}
\usepackage{sidecap}
\usepackage{amsthm}
\usepackage{amsmath}
\usepackage[square,sort,comma,numbers]{natbib}
\usepackage[linesnumbered, ruled]{algorithm2e}

\usepackage{xcolor}
\usepackage{mathtools}
\input{math_commands.tex}

\usepackage{wrapfig}
\usepackage[T1]{fontenc}
\usepackage{booktabs}
\usepackage{multirow}
\usepackage{subfigure}
\usepackage{subcaption}
\usepackage[ruled,linesnumbered]{algorithm2e}
\usepackage{enumitem}

\theoremstyle{definition}
\newtheorem{definition}{Definition}[section]

\newboolean{include-notes}
\setboolean{include-notes}{true}
\newcommand{\nb}[3]{\ifthenelse{\boolean{include-notes}}{{\colorbox{#2}{\bfseries\sffamily\scriptsize\textcolor{white}{#1}}}{\ \textcolor{#2}{\sf\small\textit{#3}}}}{}}

\begin{document}
\title{Mastering Continual Reinforcement Learning through Fine-Grained Sparse Network Allocation and Dormant Neuron Exploration}

\author{Chengqi Zheng,  Haiyan Yin, Jianda Chen, Terence Ng,  \\
Yew-Soon Ong,~\IEEEmembership{Fellow,~IEEE}, Ivor~Tsang,~\IEEEmembership{Fellow,~IEEE} 
\thanks{Chengqi Zheng, Haiyan Yin, Ivor Tsang, and Yew-Soon Ong are with the CFAR and IHPC, Agency for Science, Technology and Research (A*STAR), Singapore. Ivor Tsang and Yew-Soon Ong are also jointly affiliated with College of Computing and Data Science, Nanyang Technological University (NTU), Singapore. 
}
\thanks{Jianda Chen and Terence Ng are with the College of Computing and Data Science, Nanyang Technological University (NTU), Singapore
}
\thanks{Haiyan Yin is the corresponding author (E-mail: yin\_haiyan@cfar.a-star.edu.sg).}
}

\markboth{
}%
{Shell \MakeLowercase{\textit{et al.}}: A Sample Article Using IEEEtran.cls for IEEE Journals}


\maketitle


\begin{abstract}

Continual Reinforcement Learning (CRL) is essential for developing agents that can learn, adapt, and accumulate knowledge over time.  However, a fundamental challenge persists as agents must strike a delicate balance between \emph{plasticity}, which enables rapid skill acquisition, and \emph{stability}, which ensures long-term knowledge retention while preventing catastrophic forgetting. 
In this paper, we introduce \textbf{SSDE}, a novel structure-based approach that enhances plasticity through a fine-grained allocation strategy with \textbf{S}tructured \textbf{S}parsity and \textbf{D}ormant-guided \textbf{E}xploration. 
SSDE decomposes the parameter space into \emph{forward-transfer} (frozen) parameters and \emph{task-specific} (trainable) parameters. 
Crucially, these parameters are allocated by an efficient co-allocation scheme under sparse coding, ensuring sufficient trainable capacity for new tasks while promoting efficient forward transfer through frozen parameters. 
However, structure-based methods often suffer from rigidity due to the accumulation of non-trainable parameters, limiting exploration and adaptability. 
To address this, we further introduce a sensitivity-guided neuron reactivation mechanism that systematically identifies and resets dormant neurons, which exhibit  minimal influence in the sparse policy network during inference. This approach effectively enhance exploration while preserving structural efficiency. 
Extensive experiments on the CW10-v1 Continual World benchmark demonstrate that SSDE achieves state-of-the-art performance, reaching a \textit{success rate of 95\%}, surpassing prior methods significantly in both plasticity and stability trade-offs 
(code is available at: \textcolor{blue}{\url{https://github.com/chengqiArchy/SSDE}}).
\end{abstract}

\begin{figure}[t]
  \begin{center}
    \includegraphics[width=0.45\textwidth]{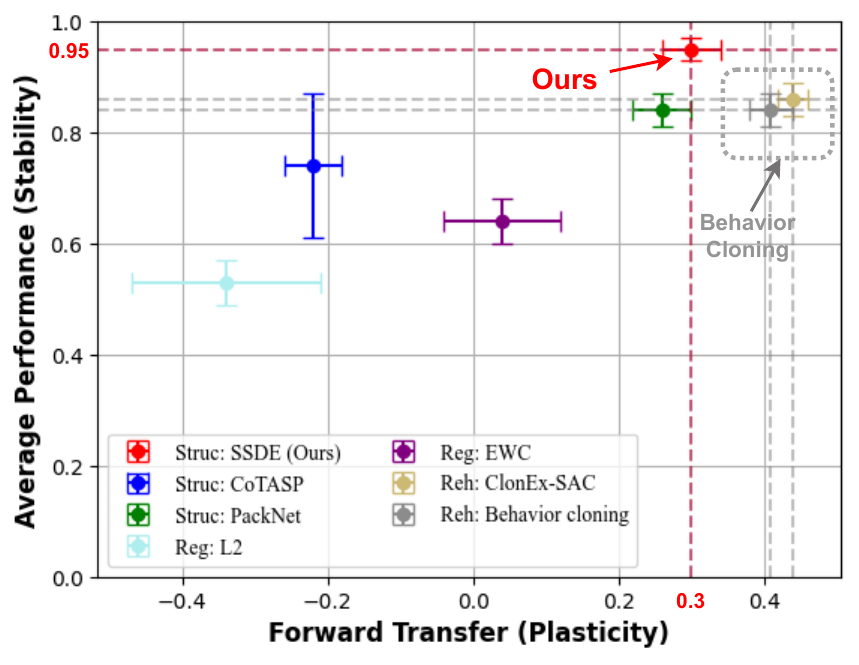}
  \end{center}
  \caption{
 \textbf{Plasticity-stability trade-off on CW10-v1 benchmark}: \emph{Stability} refers to {how well} an agent retains learned knowledge, measured by task performance; \emph{plasticity} quantifies {how quickly} an agent adapts to new tasks, measured by the normalized steps required to learn a task (Eq.~\ref{equation:metrics}). Our proposed method  \textbf{SSDE} achieves state-of-the-art \emph{stability} of 95\%, demonstrating strong capability in mitigating forgetting. For \emph{plasticity}, {SSDE} remains competitive with strong Behavior Cloning baselines, despite the latter having access to more data during experience replay. 
 }
 \vskip -0.15in
  \label{fig:stability_plasticity}
\end{figure}

\section{Introduction}
From lifelong robotics to autonomous decision-making, real-world AI systems must continuously learn without forgetting past knowledge. While humans exhibit remarkable abilities to transfer knowledge across tasks and retain learned skills, AI models, particularly reinforcement learning (RL) agents, struggle in non-stationary environments, where past knowledge can be easily overwritten or lost
~\citep{Thrun98,ChoiYZ99,LangeAMPJLST22}. 
Continual RL aims to address this challenge by enabling agents to learn sequentially without catastrophic forgetting~\citep{continual_world,khetarpal2022towards, mccloskey1989catastrophic,caruana1997multitask, pham2023continual}. 
However, achieving this requires striking a delicate balance between  \emph{plasticity} and \emph{stability} in learning systems.
\textit{Plasticity} allows agents to quickly adapt to new tasks, while \textit{stability} ensures that previously learned skills are retained~\citep{AbbasZM0M23,DohareHLRMS24}.

Existing approaches to continual RL have pursued three main approaches: (i) \emph{rehearsal}-based, which store and replay past experiences to reinforce prior knowledge~\citep{boschini2022class, zhou2023few};  (ii) \emph{regularization}-based, which impose constraints to prevent drastic updates that could overwrite previous learning;  (iii) \emph{structure}-based, which allocate dedicated sub-networks to different tasks within a shared parameter space~\citep{khetarpal2022towards,WangZSZ24}.  While rehearsal and regularization-based methods provide some degree of stability, they offer limited control over parameter interference, as experience replay and constrained optimization can introduce unintended disruptions to learned policies, making long-term retention unreliable. In contrast, structure-based methods explicitly define task boundaries and allocate independent sub-networks, effectively minimizing interference and mitigating catastrophic forgetting. Moreover, structure-based approaches eliminate the need for storing or replaying past task data, making them particularly suitable for scenarios with strict memory constraints or data inaccessibility.

One key property of structure-based methods is that they allocate \textit{sparse} sub-networks for different tasks within a shared parameter space, enabling efficient knowledge retention while reducing interference~\citep{000200YNJRIWD22,WangZSZ24}.  \emph{PackNet}~\citep{PACKNET} tackles this by iteratively pruning parameters after each task, retaining only the most crucial ones. \emph{CoTASP}~\citep{Yang00LS23} employs task-driven sparse binary masks to dynamically regulate layer outputs, ensuring task-specific adaptation. These masks, initialized via sparse coding and dictionary learning, are continuously refined through gradients derived from RL objectives, optimizing network specialization over time.
However, treating sub-network allocation as a unified process introduces a key limitation: \textit{as more tasks are added, the amount of trainable parameters progressively shrinks}. 
This gradual capacity reduction restricts the model’s ability to adapt, ultimately compromising plasticity, especially in complex tasks where greater capacity is essential, as suggested by \textit{scaling laws}~\citep{scaling_law}. Additionally, these methods often incur significant computational overhead, as sub-network allocation relies on costly procedures like pruning and gradient-based optimization.
To enhance the \textit{plasticity} of structure-based methods,  it is crucial to not only \textit{allocate sufficient trainable parameters} for new tasks but also \textit{effectively leverage previously trained parameters} for inference. Achieving this balance is essential for maintaining the expressiveness of sparse sub-network policies and ensuring  the long-term adaptability of continual RL models.

In this paper, we present \textbf{SSDE},  a novel method for enhancing \emph{plasticity} through ``\textit{\textbf{S}tructured \textbf{S}parsity and \textbf{D}ormant neuron-guided \textbf{E}xploration}'', designed to optimize the three core aspects of continual RL policies:
\begin{enumerate}[label=(\roman*)]
    \item \textbf{Allocation:} SSDE introduces a \textbf{fine-grained co-allocation} strategy based on sparse coding, explicitly partitioning sub-network parameters into \textit{forward-transfer} (fixed) and \textit{task-specific} (trainable) components. This ensures sufficient capacity for acquiring new skills while maximizing knowledge transfer efficiency.
    \item \textbf{Inference:} SSDE incorporates a dedicated inference function with a  novel \textbf{trade-off parameter} that dynamically balances forward-transfer and task-specific parameters. This prevents frozen parameters from overshadowing trainable ones, thereby expanding the solution space and enabling more flexible and adaptive inference strategies.
    \item \textbf{Training:} SSDE introduces a \textbf{sensitivity-guided dormant neuron reset algorithm} to enhance the expressiveness of sparse policies, which are inherently constrained in trainable capacity. By identifying and reactivating neurons that remain unresponsive to input sensitivity, SSDE mitigates the limitations of sparse sub-networks and improves their adaptability.
\end{enumerate}
Together, the strategic combination of \textit{fine-grained co-allocation} and exploration with \emph{dormant} neurons establish a robust foundation for {SSDE} to significantly enhance the \emph{plasticity}-\emph{stability} trade-off. We show {SSDE} not only achieves SOTA \textit{stability} but also achieves competitive \textit{plasticity} even when compared to strong behavior cloning baselines that benefit from data replay (Figure~\ref{fig:stability_plasticity}). 
We also show the consistency of {SSDE}'s performance across both v1 \& v2 of Continual World benchmark (Table~\ref{table:main} \& Table~\ref{table:cw10_v2}). 
A case study on co-allocated masks with structured sparsity highlights that our approach significantly improves parameter utilization while drastically reducing allocation time (Table~\ref{tab:mask_allocation_time} \& Figure~\ref{fig:free_vs_cotasp}). 
Visualizations of sub-network masks further demonstrate that structural sparsity effectively captures task similarities (Figure~\ref{figure:task_overlap}). 
Finally, a comprehensive ablation study (Table~\ref{table:ablation}) confirms that SSDE's core components are crucial for driving the success of the model.

\section{Related Works}

Continual RL,  a.k.a. lifelong RL, seeks to develop agents capable of continuously learning from a sequence of tasks without forgetting previous knowledge. 
For comprehensive surveys, we refer  readers to~\citep{khetarpal2022towards} and~\citep{sebastian-survey}, while a formal definition of continual RL agents can be found in~\citep{Abel0RPHS23}.
Existing approaches can be broadly categorized into \textit{rehearsal}-based, \textit{regularization}-based, and \textit{structure}-based methods. 
Among existing approaches, the current SOTA performance on continual learning benchmarks in Meta-World manipulation ~\citep{meta-world} is achieved by the rehearsal-based method {CloneEX-SAC}~\citep{ClonEX-SAC}. By storing past task data and policies for behavior cloning, it enables high forward transfer through intensive data replay, though at a cost of high computational overhead. In contrast, structure-based methods eliminate the need for data storage or reuse and use a single set of parameters to represent multiple policies, enabling more efficient training and inference.

While structure-based methods mitigate task interference by using sub-networks, they often pursue sparsity-driven allocation, which sacrifices capacity and hinders adaptation (i.e., \emph{plasticity}).
PackNet~\citep{PACKNET} prunes the network after each task, fine-tuning the dense policy into a sparse one by retaining the most important parameters, albeit with significant computational effort. 
HAT~\citep{HAT} employs hard attention masks, introducing a small number of trainable weights that are updated alongside the main model.
CSP~\citep{CSP} progressively expands the subspace of policies by integrating new policies into the space as anchors, provided they yield measurable performance gain.
Rewire~\citep{Rewire} employs a differentiable wiring mechanism to adaptively permute neuron connections, enhancing policy diversity and stability in non-stationary environments.

More recently, sparse prompting-based approaches have emerged, effectively bridging cross-modality task relationships with parameter allocation strategies.
TaDeLL~\citep{TaDeLL} employs coupled dictionary optimization to refine task descriptors and policy parameters, initializing new task policies as a sparse linear combinations over a shared basis. 
CoTASP~\citep{Yang00LS23} extends this by generating sparse sub-network masks through sparse encoding and dictionary learning, which are further refined via gradient optimization during RL training.
While  both works focus on leveraging task similarities for parameter allocation, they suffer from a fundamental limitation: as more tasks are introduced, sparse networks gradually lose trainable capacity, restricting the acquisition of new skills. 
Our work addresses  this issue with a novel co-allocation strategy built on sparse coding, designed to ensure efficient forward-transfer parameter allocation while simultaneously maintaining sufficient capacity for trainable parameters. Moreover, SSDE achieves \textit{fully preemptive allocation} (i.e., allocating task-specific parameters in advance without requiring iterative refinement), eliminating the need for computationally intensive dictionary learning or iterative updates used in CoTASP, significantly enhancing allocation efficiency and scalability.

Our work further integrates continual RL with the recently proposed \textit{dormant neuron phenomenon}~\citep{Dormant} to address a key question:
\textit{How can structure-based continual RL agents use their sparse sub-networks to their full potential? }
\cite {Dormant} proposes \emph{ReDo}, a mechanism that periodically resets inactive neurons in {full-scale dense} policies to restore network capacity without significantly altering  policy.  It identifies dormant neurons using a simple yet effective method based on  neuron activation scales. In  context of structure-based continual RL,  expressivity challenge is more pronounced due to sparse sub-networks, where a substantial portion of parameters are frozen, leaving only a small fraction trainable. SSDE extends the dormant neuron concept by proposing a sensitivity-guided dormant that intuitively identifies neurons unresponsive to observation changes, enhancing the sparse policy's responsiveness to crucial states.
Integrating this phenomenon into continual RL is crucial, as expressivity is directly tied to \textit{plasticity}, especially in sparse sub-networks where limited capacity hinders adaptability. To the best of our knowledge, SSDE is the first work to address expressivity limitations of sparse policy networks in continual RL.

\section{Problem Formulation}\label{sec:problem_formulation}
We consider the problem of continual RL under a task-incremental setting, where an RL agent sequentially learns a set of N distinct tasks, denoted as  $\mathcal{T}_{cl}=\{\mathcal{T}_1, ..., \mathcal{T}_N\}$, following ~\citep{continual_world, Yang00LS23}. Each task $\mathcal{T}_k$ is formulated as a Markov Decision Process (MDP), defined by: $\mathcal{T}_{k} = \langle \mathcal{S}^{(k)}, \mathcal{A}^{(k)}, p^{(k)}, \mathcal{R}^{(k)}, \gamma \rangle$, where $\mathcal{S}^{(k)}$ represents the state space, $\mathcal{A}^{(k)}$ is the action space, $p^{(k)}: \mathcal{S}^{(k)}_t \times \mathcal{A}^{(k)}_t \rightarrow \mathcal{S}^{(k)}_{t+1}$ is the transition probability function, $\mathcal{S}^{(k)}_{t}$, 
$\mathcal{R}^{(k)}: \mathcal{S}^{(k)}_{t} \times \mathcal{A}^{(k)}_{t} \rightarrow \mathbb{R}$ is a reward function,
and $\gamma \in [0, 1)$ is the discount factor. 
The goal of the agent is to train a single  RL policy $\pi_{\theta}^*$ that performs well across all tasks through sequential task interaction, 
\begin{equation}
\theta^{*}=\arg\max_{\theta}\sum_{k=1}^{\mathcal{T}}\mathbb{E}_{\pi_{\theta}}\left[\sum_{t=0}^{\infty}\gamma^{t}\mathcal{R}^{(k)}\Big(s^{(k)}_t,a^{(k)}_t\Big)\right]. \label{eq:goal_crl}
\end{equation}

Structure-based continual RL tackles the challenge of \emph{plasticity-stability} trade-off by dynamically partitioning the policy network into task-specific sub-networks, minimizing task interference and preserving the degradation of earlier behaviors.
Formally, for each task $\mathcal{T}_k$, it establishes a task-conditioned  mapping $\phi: \theta \times \mbox{task}_{id}{(k)} \rightarrow \theta_{k}$ that automatically maps the network parameters $\theta$ and task identity $\mbox{task}_{id}$ to generate a dedicated sub-network policy $\pi_{\theta_{k}}$, where $\theta_{k} \subseteq \theta$. 
Crucially, the space for the sub-network parameters can be decomposed into two parts: $\theta_k = \{\theta_{k}^{fw}, \theta_{k}^{sp}\}$, where $\theta_{k}^{fw}$ represents a group of the \emph{forward-transfer} parameters shared with previous tasks $\theta_{1},...,\theta_{k-1}$, frozen after training and used for inference only. Formally,  $\displaystyle \theta_{k}^{fw} = \Big(\bigcup_{i=1}^{k-1}\theta_{i}\Big) \bigcap \theta_{k}$, and the remaining are \emph{task-specific} parameters, updated solely for learning the current task, i.e., $\displaystyle \theta_{k}^{sp} = \theta_{k} \setminus \theta_{k}^{fw}$. 
Note that each task only updates its \emph{task-specific} parameters $\theta_{k}^{sp}$, while the \emph{forward-transfer} parameters $\theta_{k}^{fw}$ remain fixed to prevent task interference. To scale the agent's capability for handling multiple tasks, the sub-network parameters are typically sparse.

For efficient task allocation, we employ a neuron-level partitioning method to establish task boundaries. For each layer $l$, sub-networks are prompted by applying binary masks $\phi_k^{(l)}$ to the outputs of the $l$-th layer $y^{(l)}$, generating calibrated network outputs as follows: 
\begin{equation}
    \vy_k^{(l)} = \phi^{(l)}_{k} \otimes f(\vy_k^{(l-1)}; \theta_k^{(l)}),
\end{equation}
where $f(\cdot)$ is the conventional inference function for the $l$-th layer, $\theta_k^{(l)}=\{\theta_{k}^{{fw}_{(l)}}, \theta_{k}^{{sp}_{(l)}}\}$, and $\otimes$ is element-wise multiplication. The key to the structure-based approach lies in determining how to allocate $\theta_k^{fw}$ and $\theta_k^{sp}$ for each task to maximize the use of learned knowledge through $\theta_k^{fw}$ while ensuring sufficient capacity in $\theta_k^{sp}$ to capture new skills. However, existing structure-based RL methods allocate $\theta_k$ as a unified process, neglecting the intricate interplay between forward-transfer and task-specific parameters. As a result, these methods lead to a progressive decline in available \textit{trainable parameters} as more tasks are introduced, increasing network rigidity and reducing expressiveness, which limits the model’s ability to adapt to novel tasks and capture complex behaviors in continually evolving environments. 

\begin{figure*}[t]
\centering
\begin{minipage}{0.55\textwidth} 
    \centering
    \includegraphics[width=\textwidth]{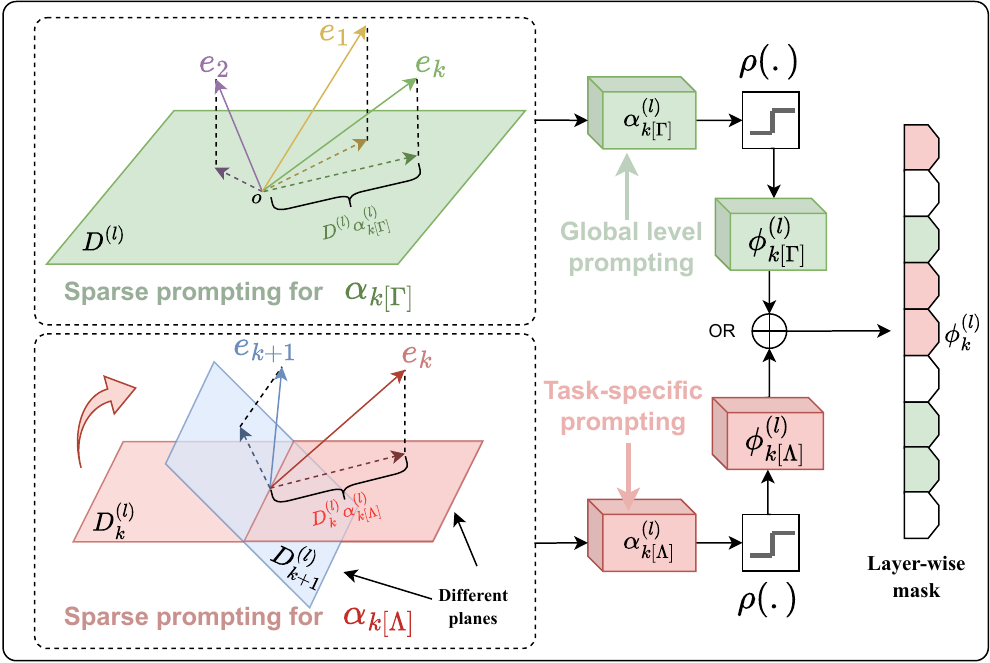} 
\end{minipage}%
\hfill 
\begin{minipage}{0.43\textwidth} 
    \centering
    \captionof{figure}{\justifying
    \textbf{Co-Allocation with Sparse Prompting} aims to learn two sets of calibration embeddings, $
    \bm{\alpha_{k[\Gamma]}}$ and $\bm{\alpha_{k[\Lambda]}}$, which generate neuron-level binary calibration masks $\bm{\phi_{k[\Gamma]}^{(l)}}$ and $\bm{\phi_{k[\Lambda]}^{(l)}}$. Both masks are further merged together to form $\bm{\phi_k^{(l)}}$, which is multiplied to the output of the $l$-th layer to calibrate the output. This layer-wise binary mask intuitively specifies a layer-wise sparse sub-net structure, promoting enhanced \emph{plasticity}.  \\
    \textbf{Upper}: A global-level sparse coding process learns $\bm{\alpha_{k[\Gamma]}}$ by projecting different task embeddings onto a shared plane of $\mD^{(l)}$, assigning similar masks to similar tasks. \\ 
    \textbf{Lower}: A local task-specific prompting process leverages random projection planes to learn $\bm{\alpha_{k[\Lambda]}}$, increasing the capacity for trainable parameters. 
    }
    \vskip -0.1in
    \label{fig:img1}
\end{minipage}
\end{figure*}

\section{Methodology}
In this section, we propose \textbf{SSDE} (\emph{plasticity} through \textbf{S}tructured \textbf{S}parsity with \textbf{D}ormant neuron-guided \textbf{E}xploration).
\begin{itemize}
\item \textbf{Sec~\ref{sec:co-allocation}: Co-Allocation with Sparse Prompting}  
We introduce a {co-allocation} algorithm that allocates parameters for each sub-network policy regarding task global relationship and local structure. The allocated parameters are decomposed into \emph{forward-transfer} (fixed) and \emph{task-specific} (trainable), allocated under an objective of preserving knowledge from previous tasks for forward transfer while maximizing the number of trainable parameters for increased network plasticity.

\item \textbf{Sec~\ref{sec:fine-grained-masking}: Fine-Grained Sub-Network Masking}
We propose a masking mechanism that enables task-specific prompting during inference and training. By applying neuron-level and parameter-level masks, sub-networks can be selectively frozen or activated, ensuring efficient parameter utilization and minimizing task interference.

\item \textbf{Sec~\ref{sec:dormant-neuron}: Sensitivity-Guided Dormant Neuron Exploration}  
To address expressiveness limitations in sparse sub-network policies with constrained trainable capacity, we propose a sensitivity-guided exploration strategy that selectively reactivates dormant neurons, enhancing adaptability and representation power.
\end{itemize}




\subsection{Fine-Grained Sub-network Allocation}
\label{sec:fg-allocation}
Sparse prompting-based approaches ~\citep{Yang00LS23,DBLP:conf/emnlp/ReimersG19} enhance plasticity by assigning sparse codes to respective tasks, which are then transformed into neuron masks to generate dedicated sub-networks. Building on this foundation, 
we propose a \textbf{sub-network co-allocation strategy} that leverages \emph{global task correlations} and  \emph{local task-specific dictionaries} for effective parameter allocation.
Specifically, we introduce \textbf{global cross-task sparse prompting}, denoted as $\alpha_{[\Gamma]}$, derived from embeddings of text descriptions encoded by pre-trained Sentence-BERT~\cite{DBLP:conf/emnlp/ReimersG19} and a global coding dictionary, to capture shared task relationships. Crucially, we also introduce \textbf{local task-specific sparse prompting}, $\alpha_{[\Lambda]}$, 
constructed from individual local dictionaries, to ensure sufficient capacity for task-specific parameters.
These two components synergistically co-allocate dedicated forward-transfer and task-specific parameters in sparse sub-networks, preventing capacity bottlenecks. 
During reinforcement learning, we
incorporate a \textbf{fine-grained masking mechanism} that flexibly freezes  \emph{forward-transfer parameters},  while selectively updating \emph{task-specific} parameters to integrate new knowledge. 

\subsubsection{\textbf{Co-Allocation with Sparse Prompting}}
\label{sec:co-allocation}

To obtain a global cross-task sparse prompting $\alpha_{k [\Gamma]}$,  we begin by generating task embeddings $e_k\in\mathbb{R}^m$ for each task $\mathcal{T}_k$ by encoding their corresponding task textual descriptions using a pre-trained language model.
For each layer-$l$ in the sub-network, we define a shared  space among all the tasks as an over-complete dictionary $\displaystyle \mD^{(l)} \in \mathbb{R}^{m\times n^{(l)}}$, where $n^{(l)}$ is the number of neurons at layer-$l$ in the full network. Elements in $\displaystyle \mD^{(l)}$ are sampled from normal distribution $\mathcal{N}(0, 1)$ and $\displaystyle \mD^{(l)}$ is fixed across all task in $\mathcal{T}_{cl}$. 
We aim to learn a sparse prompting $\alpha_{k [\Gamma]}^{(l)} \in \mathbb{R}^{n^{(l)}}$ that could reconstruct task embeddings $e_k$ as a linear combination of neuron's representations, i.e., atoms from the dictionary. 
The sparse prompting $\alpha_{k [\Gamma]}^{(l)}$ can be obtained by optimizing the Lasso problem formalized as follows,
\begin{align}
    \alpha_{k [\Gamma]}^{(l)} = \argmin_{\alpha_{k [\Gamma]}\in\mathbb{R}^{m}}\frac{1}{2}\|e_{k} - \mathbf{D}^{(l)}\alpha_{k [\Gamma]}^{(l)}\|^2_2 + \lambda_{ [\Gamma]}\|\alpha_{k [\Gamma]}^{(l)}\|_1,\quad \nonumber\\
    \text{for layer~}l=1,\dots,L-1 , \label{eq:sc_fw}
\end{align}
where $\|\cdot\|_p$ is the $L_p$ norm, $\lambda_{[\Gamma]}$ is a hyperparameter controlling the sparsity of the forward-transfer prompting $\alpha_{k [\Gamma]}^{(l)}$, and $L$ is the number of layers. A step function $\rho(\cdot)$ transforms the sparse vector into a binary mask, i.e., ${\phi}_{k [\Gamma]}^{(l)} = \rho(\alpha_{k [\Gamma]}^{(l)})$,
where ${\phi}_{k [\Gamma]}^{(l)} \in \{0,1\}^{n^{(l)}}$ determines  the active neurons for task $\mathcal{T}_k$ at layer-$l$.
Through this global-level optimization, similar tasks will share neuron masks that allocate similar neurons.
However, excessive overlap of fixed neurons across tasks reduces the number of available trainable parameters for new tasks, limiting adaptation capacity. 

To mitigate this issue, we introduce \textbf{ task-specific sparse prompting}, which projects each task $\mathcal{T}_k$ into a \textbf{unique} dictionary $\displaystyle \mathbf{D}_k^{(l)}$, reducing redundancy and ensuring more distinct task representations. Unlike the global dictionary, which captures shared structure across tasks, this task-specific dictionary is designed to increase independence in sub-network allocation, even for tasks with similar descriptions.
Inspired by sparse random projection techniques~\citep{Li_2006}, we construct task-specific sparse prompting using over-complete dictionaries sampled from $\mathcal{N}(0,1)$. 
The local task-specific prompting $\alpha_{k [\Lambda]}^{(l)}$ that selects the trainable neurons is learned by solving the following objective:
\begin{align}
    \alpha_{k [\Lambda]}^{(l)} = \argmin_{\alpha_{k [\Lambda]}\in\mathbb{R}^{m}}\frac{1}{2}\|e_{k} - \mathbf{D}_k^{(l)}\alpha_{k [\Lambda]}^{(l)}\|^2_2 +
    \lambda_{ [\Lambda]}\|\alpha_{k [\Lambda]}^{(l)}\|_1, 
    \nonumber\\ \quad\text{for layer~}l=1,\dots,L-1, \label{eq:sc_free}
\end{align}
where $\lambda_{[\Lambda]}$ controls the sparsity level of $\alpha_{k [\Lambda]}^{(l)}$. 
To efficiently compute the sparse prompting vectors, we employ a Cholesky-based implementation of the LARS algorithm~\citep{Efron_2004}, which offers a good balance of performance and ease of implementation.
The final task-specific binary mask is obtained using a step function:  $\phi_{k [\Lambda]}^{(l)} = \rho(\alpha_{k [\Lambda]}^{(l)})$, where $\rho(\cdot)$ is a threshold function.

The final co-allocated sub-task masks, $\phi_{k}^{(l)}$, are obtained by combining the two groups of masks: $\phi_{k}^{(l)} = \phi_{k[\Gamma]}^{(l)} \vee \phi_{k[\Lambda]}^{(l)}$, where $\vee$ denotes the element-wise OR operation, and each element in the mask is a Boolean value (0 or 1). This co-allocation with sparse prompting process ensures that each task's sub-network includes high-quality forward-transfer parameters for rapid adaptation, while preserving sufficient trainable capacity for task-specific learning. Furthermore, this mask learning process is computationally efficient and independent of RL interactions, as it relies solely on task description embeddings rather than gradient-based optimization.

\subsubsection{\textbf{Fine-Grained Sub-Network Masking}}
\label{sec:fine-grained-masking}
With the neuron-level mask $\phi_k^{(l)}$, we can determine which parameters in weight matrix $W^{(l)}\in \mathbb{R}^{n^{(l)} \times n^{(l-1)}}$ are actively used in task $\mathcal{T}_k$, effectively determining the allocated sub-network. We denote the \textbf{binary mask matrix}  as $\tilde{\Psi}_k^{(l)} \in \mathbb{R}^{n^{(l)} \times n^{(l-1)}}$,    computed through matrix-multiplication of $\phi_k^{(l)}$ with the previous layer's neuron mask $\phi_k^{(l-1)}$:
\begin{equation}
\label{eq:Psi}
    \tilde{\Psi}_k^{(l)} = \phi_k^{(l)}  (\phi_k^{(l-1)})^T ,
\end{equation}
where an element is set to  1 if the parameter is activated in task $\mathcal{T}_k$. 
Specially, $\phi_k^{(0)} = \mathbf{1}$ and $\phi_k^{(L)} =  \mathbf{1}$ are vectors where all elements are ones. 
The element at row $p$ and column $q$ in matrix $\tilde{\Psi}_k^{(l)}$ is one if and only if the $p$-th element of $\phi_k^{(l)}$ and $q$-th element of $\phi_k^{(l-1)}$ are both ones.
Therefore, by construction, $\tilde{\Psi}_k^{(l)}$ is at least as sparse as  $\phi_k^{(l)}$ and $\phi_k^{(l-1)}$. 
As mentioned in Section~\ref{sec:problem_formulation}, the parameters selected by $\tilde{\Psi}_k^{(l)}$ are allocated as the \textbf{sub-network parameters} for the current task $\mathcal{T}_{k}$ and a subset of them is retained as  \emph{forward-transfer}  parameters $\theta^{fw}$ frozen for future tasks $\mathcal{T}_{k+1:N}$ to prevent catastrophic forgetting.

\begin{figure*}[t]
\centering
\begin{minipage}{0.58\textwidth} 
    \centering
    \includegraphics[width=\textwidth]{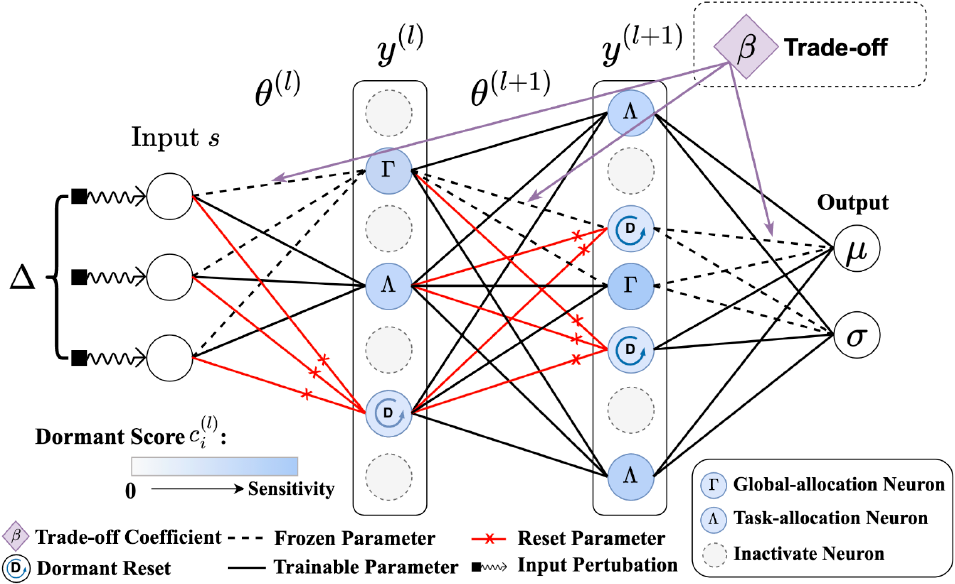} 
\end{minipage}%
\hfill 
\begin{minipage}{0.38\textwidth}
    \centering
    \captionof{figure}{\justifying
    \textbf{Structural Exploration with Dormant Neurons in SSDE}: 
  (i) {Structural sparsity} is achieved by generating a sub-network from neurons \textbf{co-allocated by two sparse prompting processes} ($\Gamma$ and $\Lambda$).  (ii) Fine-grained inference is performed on it, with the \textbf{trade-off coefficient {$\boldsymbol{\beta}$}} controlling the balance of \textit{trainable} and \textit{frozen} parameters.
   (iii) For structural exploration, the input of the sparse network is perturbed to maximize the sensitivity of active neurons. 
  Neurons colored blue (acquired for sub-network policy during  co-allocation) are evaluated based on sensitivity score $c_i^{(l)}$, while inactive neurons, identified as \textit{\textbf{dormant}} (marked \textcolor{black}{`D'}) are reset to enhance sub-network expressiveness.
  }
\label{figure:sensitivity-guided_exploration}
\end{minipage}
\end{figure*}

Many prior structure-based methods~\citep{PACKNET,Yang00LS23} freeze parameters at the neuron-level, meaning entire rows in $W^{(l)}$ are fixed after training task $\mathcal{T}_k$. However, this approach \textbf{wastes} many parameters, particularly those selected by $\phi_k^{(l)}$ but not covered by $\phi_k^{(l-1)}$, as they remain untrained throughout the process. This severely restricts network plasticity, progressively reducing the capacity for new task adaptation.
To address this limitation, our work proposes \textbf{fine-grained sub-network masking} mechanism that \textbf{freezes  only the exact parameters trained in previous tasks}  $\mathcal{T}_{1:k-1}$, while keeping others available for future learning. To this end, we maintain a mask matrix ${\Psi}_{k-1}^{(l)}$ that tracks frozen parameters, formally defined as performing element-wise OR operation across all seen fine-grained masks $\tilde{\Psi}_i^{(l)}$:
\begin{equation} \label{eq:frozen_mask} 
{\Psi}_{k-1}^{(l)} = \vee_{i=1}^{k-1}\tilde{\Psi}^{(l)}_{i}, \quad\text{for task~}i=1,\dots,k-1.  \end{equation}

The fine-grained masking mechanism precisely  identifies the subset of parameters, ${\Psi}_{k-1}^{(l)}$ that were utilized in previous tasks $\mathcal{T}_{1:k-1}$ and should be frozen  starting from task $\mathcal{T}_k$. Building on this, we formulate a fine-grained inference function for each layer, effectively decomposing the forward pass into two distinct components: 
one for  \emph{task-specific} (trainable) parameters, and another for  \emph{forward-transfer} (frozen) parameters: 
\begin{equation}    
\label{eq:calibration}
\begin{split}
   \hat{\vy}^{(l)}_k  
     &= \underbrace{\left((1-\Psi_{k-1}^{(l)}) \otimes \tilde{\Psi}_{k}^{(l)} \otimes \mW^{(l)}\right)}_{\text{ task-specific parameters (trainable)}} \vy_k^{(l-1)} + \\
     &\underbrace{{\beta}\vphantom{\left((1-\Psi_{k-1}^{(l)}) \otimes \tilde{\Psi}_{k}^{(l)} \otimes \mW^{(l)}\right)} }_{\text{trade-off}} \cdot\underbrace{\left(\Psi_{k-1}^{(l)}\otimes \tilde{\Psi}_{k}^{(l)} \otimes W^{(l)}\right)}_{\text{ forward-transfer parameters (frozen)}} \vy_k^{(l-1)}   + b_k^{(l)} \otimes  \phi_k^{(l)},
\end{split}
\end{equation}
where $\hat{\vy}^{(l)}_k$ is the pre-activation output, and the  layer output is $\vy^{(l)}_k=h(\hat{\vy}^{(l)}_k)$, with $h(\cdot)$ as the activation function. A critical component of this formulation is \textbf{a novel trade-off parameter} $\beta$, which provides fine-grained control over the influence of forward-transfer parameters. As the task distribution evolves, the capacity of frozen parameters increases while the availability of trainable parameters decreases. Our fine-grained inference mechanism dynamically adjusts  $\beta$ to control the balance of pre-trained knowledge with the acquisition of new skills, preventing the pre-trained knowledge from overshadowing the trainable parameters, enhancing the plasticity, and enabling the model learn new tasks effectively.  
Figure~\ref{figure:sensitivity-guided_exploration} shows an example of the fine-grained inference procedure in SSDE.

\noindent\textbf{Learning:} 
To optimize our proposed fine-grained sub-network allocation method, we update only the task-specific parameters using masked gradient descent:
\begin{equation}\label{eq:sgd_update}
\centering
    \theta^{(l)} \leftarrow \theta^{(l)} - \alpha\,(1-{\Psi}_{k-1}^{(l)}) \otimes \vg_k^{(l)},
\end{equation}
where $\alpha$ is the learning rate, and $\vg_k^{(l)}$ is the gradient w.r.t.  parameters $\theta^{(l)}$.
Parameters corresponding to ${\Psi}_{k-1}^{(l)} = 1$ remain frozen, i.e., their gradients are set to zero. This effectively stops gradient updates for the forward-transfer term $\Psi_{k-1}^{(l)}\otimes \tilde{\Psi}_{k}^{(l)} \otimes W^{(l)}$ with respect to $W^{(l)}$, ensuring previously learned knowledge remains intact.
A detailed version of the co-allocation method is provided in Algorithm~\ref{algo}.


\subsection{Sensitivity-Guided Dormant Neuron Reset}
\label{sec:dormant-neuron}

Training sparse prompted sub-network policies in the continual RL  often encounters a fundamental challenge: \textbf{limited expressivity due to the increasing \emph{rigidity} of the policy network over tasks}. As training progresses, the proportion of non-trainable parameters grows, dominating the network's output and restricting its adaptability to new tasks. 
While freezing parameters helps prevent catastrophic forgetting, it also reduces the availability of trainable parameters, leading to a \textbf{loss of plasticity} and restricting the network’s ability to adapt to new tasks. Consequently, only a subset of neurons remains active, and the policy becomes increasingly deterministic, limiting its exploration and expressiveness.

To enhance the adaptability of sparse sub-networks, we propose a novel \textbf{sensitivity-guided structural exploration} strategy facilitated by a newly defined \textbf{sensitivity-guided dormant score}. Inspired by the dormant neurons phenomenon~\citep{Dormant}, our approach involves \textbf{periodically resetting} neurons that have become unresponsive. Unlike  prior work, which evaluates neuron responsiveness solely by neuron activation magnitude in full-scale networks, SSDE accesses a neuron's sensitivity to input variations, a crucial aspect in sparse sub-networks where limited trainable parameters hinder exploration and learning.
Appendix\ref{apendix:case_study_T5} provides further insights into this phenomenon.

\begin{definition}[Sensitivity-Guided Dormant Score]
Let $\vy_{k,(i)}^{(l)}(\vs)$ denote the $i$-th neuron output of layer-$l$ given observation $\vs$ as the input, and $ \Delta$ be noise vector to perturb $\vs$. Given a observation distribution $\mathcal{D}_{\vs}$ and $\vs \in \mathcal{D}_{\vs}$, the sensitive-dormant score of neuron $i$ at layer-$l$ is defined as:

\begin{equation}\label{eq:dormant_sensitivity}
    c_i^{(l)} = \frac{\mathbb{E}_{\vs \in\mathcal{D}_{\vs}} \left|\vy_{k,(i)}^{(l)}(\vs) - \vy_{k,(i)}^{(l)}(\vs+ \Delta)\right|}{\frac{1}{\left\|\phi_k^{(l)}\right\|_1} \sum_{j}\mathbb{E}_{\vs \in\mathcal{D}_{\vs}}\left|\vy_{k,(j)}^{(l)}(\vs) - \vy_{k,(j)}^{(l)}(\vs+ \Delta)\right|}. \\
\end{equation}

We say a neuron $i$ in layer $l$ is $\tau$-\textbf{dormant} if $c_i^{(l)} \leq \tau$. 
\end{definition}

The sensitivity-guided dormant score offers a quantitative measure of parameter rigidity in sparse sub-networks, enabling targeted reactivation of underutilized neurons.
Formally, we introduce a reset mechanism that involves injecting controlled perturbation noise into input observations and analyzing output variations across the sub-network layers. 
Neurons exhibiting significant output changes are retained for structural exploration, while unresponsive neurons are reset to restore expressivity and enhance plasticity, enabling better adaptation to new skills in structure-based continual RL.
At the beginning of training, we store the randomly initialized values of all parameters. 
We periodically evaluate the sensitivity-guided dormant scores for all neurons at fixed training intervals where the scores are computed according to Equation~\ref{eq:dormant_sensitivity}.
As illustrated in Figure~\ref{figure:sensitivity-guided_exploration}, neurons with scores $c_i^{(l)} \leq \tau$ are designated as dormant. Only the \textbf{trainable} \emph{task-specific} parameters connected to these dormant neurons are \textbf{reset} to their initial stored values. In contrast, all \textbf{frozen} parameters are maintained \textbf{unchanged}, irrespective of their connection with dormant neurons. 
\begin{algorithm}[t]
\caption{SSDE: Structured Sparsity with Dormant Neuron-Guided Exploration}
\label{algo}
\KwIn{Task sequence $\mathcal{T} = \{T_1, \dots, T_N\}$, Max training steps $T$, Dormant threshold $\tau$}
\KwOut{Trained continual RL policy $\pi$}
\textbf{Initialize} policy $\pi$, replay buffer $\mathcal{B}$, co-allocation masks $\Phi$, and dormant neurons $\mathcal{D}$;\\
\ForEach{task $T_k$ in $\mathcal{T}$}{
    \textcolor{blue}{/* Sparse Prompting-based Co-Allocation  */}\\
    Compute task embedding $e_k$ using Sentence-BERT; \\
    Optimize global sparse prompting $\alpha_k[\Gamma]$ via Lasso; \\
    Optimize task-specific sparse prompting $\alpha_k[\Lambda]$; \\
    Compute sub-network mask $\Phi_k = \rho(\alpha_k[\Gamma]) \vee \rho(\alpha_k[\Lambda])$; \\
    \textcolor{blue}{/* Train RL policy on allocated sub-network */} \\
    \For{$t = 1$ to $T$}{
        Observe state $s_t$ and select action $a_t \sim \pi(a_t | s_t, \Phi_k)$; \\
        Execute action, receive $(s_{t+1}, r_t)$, store $(s_t, a_t, r_t, s_{t+1})$ in $\mathcal{B}$; \\
        Update policy $\pi$ using masked gradient updates; \\

        \textcolor{blue}{/* Sensitivity-Guided Dormant Neuron Reset */}\\
        \If{$t \mod$ reset\_interval == 0}{
            Compute sensitivity scores $c_i^{(l)}$ for each neuron;\\
            Identify dormant neurons $\mathcal{D} = \{ i \mid c_i^{(l)} \leq \tau \}$;\\
            Reset task-specific parameters of $\mathcal{D}$ to initialization;\\
        }
    }
    \textcolor{blue}{/* Consolidate Parameters for New Tasks */}\\
    Update $\Phi_k^{\text{frozen}} = \Phi_k$ for future tasks; \\
}
\Return Trained continual RL policy $\pi$; \\
\end{algorithm}

\leavevmode
\section{Experiments}

\subsection{Experimental Settings}
\vskip 0.05in
\noindent{\textbf{Benchmarks:}} To assess the performance of SSDE, we follow the standard Continual World experimental setup from~\citep{ClonEX-SAC} and conduct extensive evaluations. Our primary benchmark is CW10 from Continual World~\citep{continual_world}, which features 10 representative manipulation tasks drawn from Meta-World~\citep{meta-world}. Additionally, we also use CW20, a version of CW10 repeated twice, to evaluate the transferability of the learned policy across repeated tasks\footnote{
\url{https://sites.google.com/view/continualworld}.
}.

\vskip 0.05in
\noindent{\textbf{Evaluation Metrics:}} We employ three key metrics, as introduced by~\cite{continual_world}:
\begin{enumerate}
    \item \textit{Average Performance} ($P$) ($\uparrow$): the average performance for all tasks, $P(t) = \frac{1}{|\mathcal{T}_{cl}|} \sum_{k=1}^{|\mathcal{T}_{cl}|} p_k(t)$, where $p_k(t)$ is the success ratio of the $k$-th task at step $t$.
    \item \textit{Forgetting} ($F$) ($\downarrow$): the average loss in performance across all tasks after learning is complete, $F = \frac{1}{|\mathcal{T}_{cl}|} \sum_{k=1}^{|\mathcal{T}_{cl}|} \left[ p_k(k \cdot \delta) - p_k(|\mathcal{T}_{cl}| \cdot \delta) \right]$, where $\delta$ represents the number of environment steps allocated for each task. 
    \item \textit{Forward Transfer} ($FT$) ($\uparrow$): the transfer is computed as a normalized area between the training curve of the compared method and the training curve of a single-task reference method trained from scratch (no adaptation). The reference performance is denoted as $p_k^b \in [0, 1]$, and the forward transfer is measured as:
\begin{align}
\label{equation:metrics}
FT_k := \frac{\text{AUC}_k - \text{AUC}_k^b}{1 - \text{AUC}_k^b}, 
\text{AUC}_k &:= \frac{1}{\delta} \int_{(k-1) \cdot \delta}^{k \cdot \delta} p_k(t) \, dt,  \nonumber \\ 
\text{AUC}_k^b &:= \frac{1}{\delta} \int_0^{\delta} p_k^b(t) \, dt. 
\end{align}  
\end{enumerate}

\vskip 0.05in
\noindent{\textbf{Training Details:}} To ensure the reliability and comparability of our experiments, we follow the training details outlined in~\citep{continual_world}, implementing all baseline methods using Soft Actor-Critic (SAC)~\citep{SAC}. 
To ensure a fair comparison across tasks, we limit the number of environment interaction steps to 1e6 per task, with each result averaged over five random seeds. And the $\Delta$ is defined as 0.01 times the average state over the preceding 1,000 steps.
Additional implementation details for SSDE are presented in Appendix\ref{appendix:implementation}.

\subsection{Evaluation of Sub-Network Co-Allocation}

\begin{table*}[t!]
\addtolength{\tabcolsep}{-1pt}
\small
\centering
\caption{Benchmark evaluation results on Continual World (v1).}
\label{table:main}
\resizebox{0.8\linewidth}{!}{
\begin{tabular}{c@{\hspace{5pt}}lr@{\hspace{-1pt}}lr@{\hspace{-1pt}}lr@{\hspace{-1pt}}l|r@{\hspace{-1pt}}lr@{\hspace{-1pt}}lr@{\hspace{-1pt}}l}
\toprule
\multicolumn{2}{l}{\textbf{Benchmarks-v1}} & \multicolumn{6}{c}{\textbf{CW 10}} & \multicolumn{6}{c}{\textbf{CW 20}} \\
\midrule
\multicolumn{2}{l}{Metrics} & \multicolumn{2}{c}{$P$ ($\uparrow$)} & \multicolumn{2}{c}{$F$ ($\downarrow$)} & \multicolumn{2}{c|}{$FT$ ($\uparrow$)} & \multicolumn{2}{c}{$P$ ($\uparrow$)} & \multicolumn{2}{c}{$F$ ($\downarrow$)} & \multicolumn{2}{c}{$FT$ ($\uparrow$)} \\
\midrule
\multirow{5}{*}{\rotatebox[origin=c]{90}{\shortstack{Reg}}} 
&\multicolumn{1}{l}{L2~\citep{L2}} 
&\colorbox{white}{0.42}&{\color[HTML]{525252}$\pm$0.10} 
&\colorbox{white}{0.02}&{\color[HTML]{525252}$\pm$0.02} 
&\colorbox{white}{-0.57}&{\color[HTML]{525252}$\pm$0.20} 
&\colorbox{white}{0.43}&{\color[HTML]{525252}$\pm$0.04} 
&\colorbox{white}{0.02}&{\color[HTML]{525252}$\pm$0.01} 
&\colorbox{white}{-0.71}&{\color[HTML]{525252}$\pm$0.10}  \\
&\multicolumn{1}{l}{EWC~\citep{EWC}} 
&\colorbox{white}{0.66}&{\color[HTML]{525252}$\pm$0.05} 
&\colorbox{white}{0.03}&{\color[HTML]{525252}$\pm$0.02} 
&\colorbox{white}{0.05}&{\color[HTML]{525252}$\pm$0.07} 
&\colorbox{white}{0.60}&{\color[HTML]{525252}$\pm$0.03} 
&\colorbox{white}{0.03}&{\color[HTML]{525252}$\pm$0.03} 
&\colorbox{white}{-0.17}&{\color[HTML]{525252}$\pm$0.07} \\
&\multicolumn{1}{l}{MAS~\citep{MAS}} 
&\colorbox{white}{0.59}&{\color[HTML]{525252}$\pm$0.03} 
&\colorbox{white}{-0.02}&{\color[HTML]{525252}$\pm$0.01} 
&\colorbox{white}{-0.35}&{\color[HTML]{525252}$\pm$0.07} 
&\colorbox{white}{0.50}&{\color[HTML]{525252}$\pm$0.02} 
&\colorbox{white}{0.00}&{\color[HTML]{525252}$\pm$0.01} 
&\colorbox{white}{-0.52}&{\color[HTML]{525252}$\pm$0.05}  \\
&\multicolumn{1}{l}{VCL~\citep{VCL}} 
&\colorbox{white}{0.58}&{\color[HTML]{525252}$\pm$0.04} 
&\colorbox{white}{-0.02}&{\color[HTML]{525252}$\pm$0.01} 
&\colorbox{white}{-0.43}&{\color[HTML]{525252}$\pm$0.13} 
&\colorbox{white}{0.47}&{\color[HTML]{525252}$\pm$0.02} 
&\colorbox{white}{0.01}&{\color[HTML]{525252}$\pm$0.02} 
&\colorbox{white}{-0.48}&{\color[HTML]{525252}$\pm$0.08}  \\
&\multicolumn{1}{l}{Fine-tuning} 
&\colorbox{white}{0.12}&{\color[HTML]{525252}$\pm$0.00} 
&\colorbox{white}{0.72}&{\color[HTML]{525252}$\pm$0.02} 
&\colorbox{white}{0.32}&{\color[HTML]{525252}$\pm$0.03} 
&\colorbox{white}{0.05}&{\color[HTML]{525252}$\pm$0.00} 
&\colorbox{white}{0.72}&{\color[HTML]{525252}$\pm$0.03} 
&\colorbox{white}{0.20}&{\color[HTML]{525252}$\pm$0.03} \\
\midrule
\multirow{4}{*}{\rotatebox[origin=c]{90}{\shortstack{Struc}}} 
&\multicolumn{1}{l}{PackNet~\citep{PACKNET}} 
&\colorbox{white}{0.83}&{\color[HTML]{525252}$\pm$0.04} 
&\colorbox{white}{0.00}&{\color[HTML]{525252}$\pm$0.00} 
&\colorbox{white}{0.21}&{\color[HTML]{525252}$\pm$0.05} 
&\colorbox{white}{0.80}&{\color[HTML]{525252}$\pm$0.01} 
&\colorbox{white}{0.00}&{\color[HTML]{525252}$\pm$0.00} 
&\colorbox{white}{0.18}&{\color[HTML]{525252}$\pm$0.03}  \\
&\multicolumn{1}{l}{HAT~\citep{HAT}} 
&\colorbox{white}{0.68}&{\color[HTML]{525252}$\pm$0.12} 
&\colorbox{white}{0.00}&{\color[HTML]{525252}$\pm$0.00} 
&\colorbox{white}{}&{\color[HTML]{525252}$-$} 
&\colorbox{white}{0.67}&{\color[HTML]{525252}$\pm$0.08} 
&\colorbox{white}{0.00}&{\color[HTML]{525252}$\pm$0.00} 
&\colorbox{white}{}&{\color[HTML]{525252}$-$}  \\
%
%
&\multicolumn{1}{l}{CoTASP\footnotemark[2]~\citep{Yang00LS23}}
&\colorbox{white}{0.73}&{\color[HTML]{525252}$\pm$0.11} 
&\colorbox{white}{0.00}&{\color[HTML]{525252}$\pm$0.00} 
&\colorbox{white}{-0.21}&{\color[HTML]{525252}$\pm$0.04}
&\colorbox{white}{0.74}&{\color[HTML]{525252}$\pm$0.03} 
&\colorbox{white}{0.00}&{\color[HTML]{525252}$\pm$0.01} 
&\colorbox{white}{-0.19}&{\color[HTML]{525252}$\pm$0.02}  \\
\midrule
\multirow{3}{*}{\rotatebox[origin=c]{90}{\shortstack{Reh}}} 
&\multicolumn{1}{l}{Reservoir} 
&\colorbox{white}{0.29}&{\color[HTML]{525252}$\pm$0.03} 
&\colorbox{white}{0.03}&{\color[HTML]{525252}$\pm$0.01} 
&\colorbox{white}{-1.11}&{\color[HTML]{525252}$\pm$0.08} 
&\colorbox{white}{0.12}&{\color[HTML]{525252}$\pm$0.03} 
&\colorbox{white}{0.07}&{\color[HTML]{525252}$\pm$0.02} 
&\colorbox{white}{-1.33}&{\color[HTML]{525252}$\pm$0.08}  \\
&\multicolumn{1}{l}{A-GEM~\citep{A-GEM}} 
&\colorbox{white}{0.14}&{\color[HTML]{525252}$\pm$0.05} 
&\colorbox{white}{0.73}&{\color[HTML]{525252}$\pm$0.01} 
&\colorbox{white}{0.28}&{\color[HTML]{525252}$\pm$0.01} 
&\colorbox{white}{0.07}&{\color[HTML]{525252}$\pm$0.02} 
&\colorbox{white}{0.70}&{\color[HTML]{525252}$\pm$0.01} 
&\colorbox{white}{0.13}&{\color[HTML]{525252}$\pm$0.03}  \\
&\multicolumn{1}{l}{ClonEx-SAC~\citep{ClonEX-SAC}} 
&\colorbox{white}{\underline{0.86}}&{\color[HTML]{525252}$\pm$0.02} 
&\colorbox{white}{0.02}&{\color[HTML]{525252}$\pm$0.02} 
&\colorbox{white}{0.44}&{\color[HTML]{525252}$\pm$0.02} 
&\colorbox{white}{\textbf{\underline{0.87}}}&{\color[HTML]{525252}$\pm$0.01}
&\colorbox{white}{0.02}&{\color[HTML]{525252}$\pm$0.01}
&\colorbox{white}{0.54}&{\color[HTML]{525252}$\pm$0.02}  \\
\midrule
\multirow{2}{*}{\rotatebox[origin=c]{90}{\shortstack{MT}}} 
&\multicolumn{1}{l}{MTL~\citep{meta-world}} 
&\colorbox{white}{0.51}&{\color[HTML]{525252}$\pm$0.10} 
&\colorbox{white}{}&{\color[HTML]{525252}$-$} 
&\colorbox{white}{}&{\color[HTML]{525252}$-$} 
&\colorbox{white}{0.51}&{\color[HTML]{525252}$\pm$0.11} 
&\colorbox{white}{}&{\color[HTML]{525252}$-$} 
&\colorbox{white}{}&{\color[HTML]{525252}$-$}  \\
&\multicolumn{1}{l}{MTL+PopArt~\citep{DBLP:conf/aaai/HesselSE0SH19}} 
&\colorbox{white}{0.66}&{\color[HTML]{525252}$\pm$0.04} 
&\colorbox{white}{}&{\color[HTML]{525252}$-$} 
&\colorbox{white}{}&{\color[HTML]{525252}$-$} 
&\colorbox{white}{0.65}&{\color[HTML]{525252}$\pm$0.03} 
&\colorbox{white}{}&{\color[HTML]{525252}$-$} 
&\colorbox{white}{}&{\color[HTML]{525252}$-$}  \\
\midrule
&\multicolumn{1}{l}{\textbf{SSDE (Ours)}} 
&\colorbox{white}{\textbf{0.95}}&{\color[HTML]{525252}$\pm$0.02} 
&\colorbox{white}{0.00}&{\color[HTML]{525252}$\pm$0.00} 
&\colorbox{white}{0.30}&{\color[HTML]{525252}$\pm$0.02} 
&\colorbox{white}{\textbf{0.87}}&{\color[HTML]{525252}$\pm$0.02} 
&\colorbox{white}{0.00}&{\color[HTML]{525252}$\pm$0.00} 
&\colorbox{white}{0.29}&{\color[HTML]{525252}$\pm$0.02}  \\
\bottomrule
\end{tabular}
}
\end{table*}
\footnotetext[2]{Reproduced from \url{https://github.com/stevenyangyj/CoTASP}. 
}

\begin{figure}[t!]
  \begin{center}
    \subfigure[Network Utilization]{
      \includegraphics[width=0.33\textwidth]{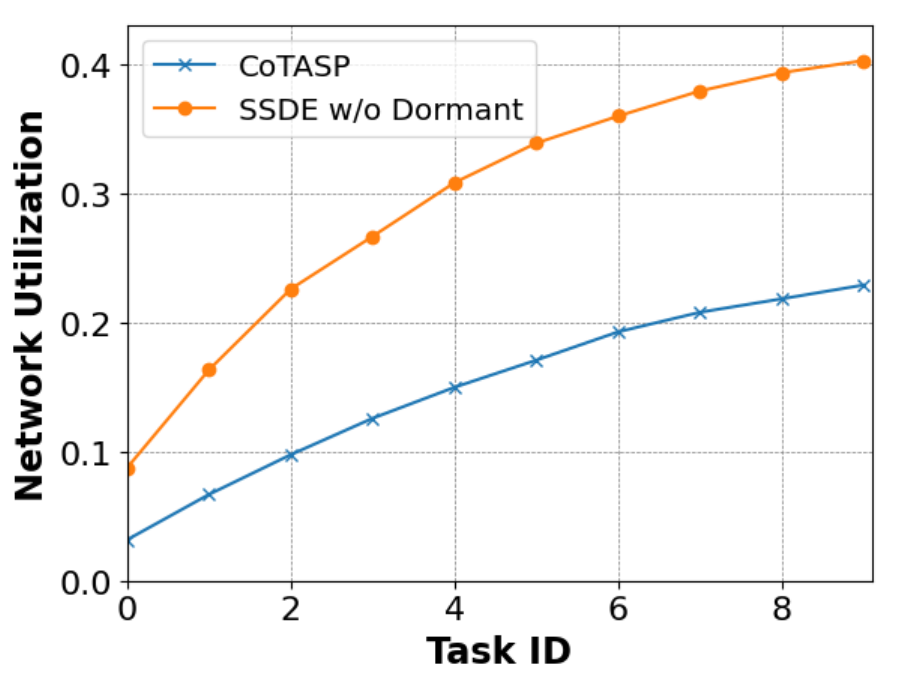}
    }
    \subfigure[Average Task Performance on CW10-v1.]{
      \includegraphics[width=0.33\textwidth]{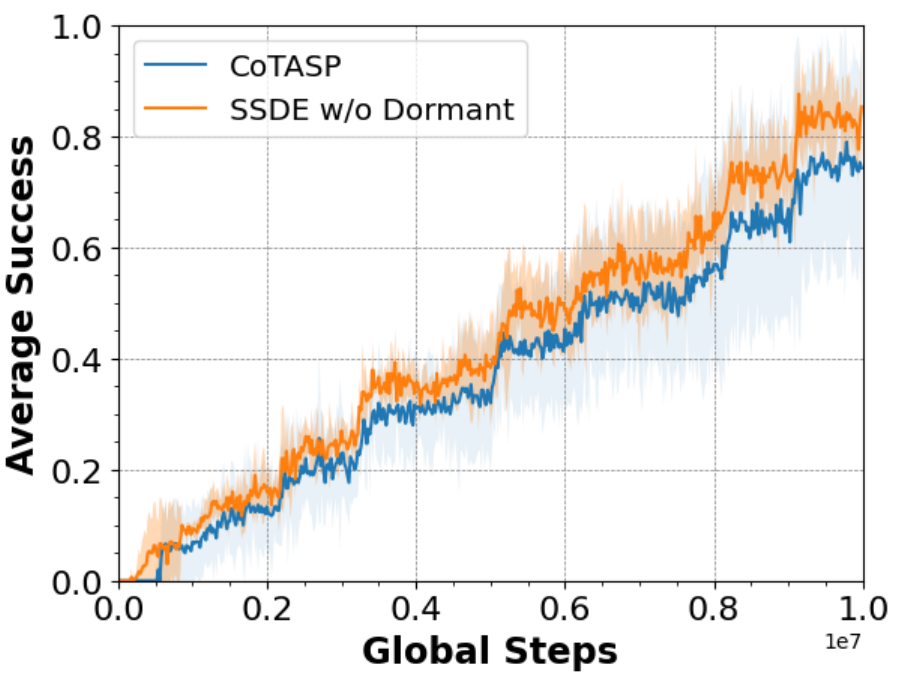}
    }
  \end{center}
  \vskip -0.1in
  \caption{Evaluation on SSDE's co-allocation vs. CoTASP's sparse prompting on CW10-v1.}
  \vskip -0.1in
  \label{fig:free_vs_cotasp}
\end{figure}

We begin with a proof-of-concept experiment to demonstrate the advantage of our proposed network allocation strategy, \textbf{sparse prompting with fine-grained co-allocation}, over the sparse prompting in CoTASP. SSDE’s co-allocation not only captures task similarity for high-quality $\theta^{fw}$ but also ensures an adequate allocation of task-specific $\theta^{sp}$ to effectively learn new knowledge, significantly enhancing the expressivity of the sparse network.
Figure~\ref{fig:free_vs_cotasp} illustrates the network utilization ratio under our method and CoTASP, measured as $\frac{\#\text{trained\_parameters}}{\text{\#total\_parameters}}$.
Our method achieves a much higher utilization ratio, using nearly $40\%$ of parameters compared to CoTASP’s $<25\%$, reducing parameter waste. Additionally, SSDE consistently outperforms CoTASP in success ratio, highlighting that co-allocation generates sub-networks with greater capacity, leading to improved \emph{plasticity}. 

\begin{table}[h]
\centering
\caption{Allocation efficiency.}
\label{tab:mask_allocation_time}
\scalebox{1.25}{
\begin{tabular}{lll}
\toprule
\textbf{Method} & \textbf{Allocation Time $(\downarrow)$} \\
\midrule
CoTASP  & 72.2s (6.45$\times$) \\
PackNet & 422.0s (37.68$\times$) \\
\textbf{SSDE (Ours)} & \textbf{11.2s (1$\times$)} \\
\bottomrule
\end{tabular}
}
\end{table}

We also examine the computational efficiency of our method compared to its closest structure-based counterparts, PackNet and CoTASP. Table \ref{tab:mask_allocation_time} reports the per-task sub-network allocation time. The overhead for PackNet is due to its computationally intensive network pruning (i.e., fine-tuning process with a significant amount of data) and that for CoTASP stems from dictionary learning and gradient-based optimization. As a result, PackNet requires more than 37$\times$ over SSDE, and CoTASP takes more than 6$\times$.  These results highlight that SSDE generates high-quality sub-networks with significantly greater computational efficiency.

\begin{figure*}[t!]
\centering
    \subfigure[Task Description $e_k$]
    {
        \includegraphics[width=0.3\linewidth]{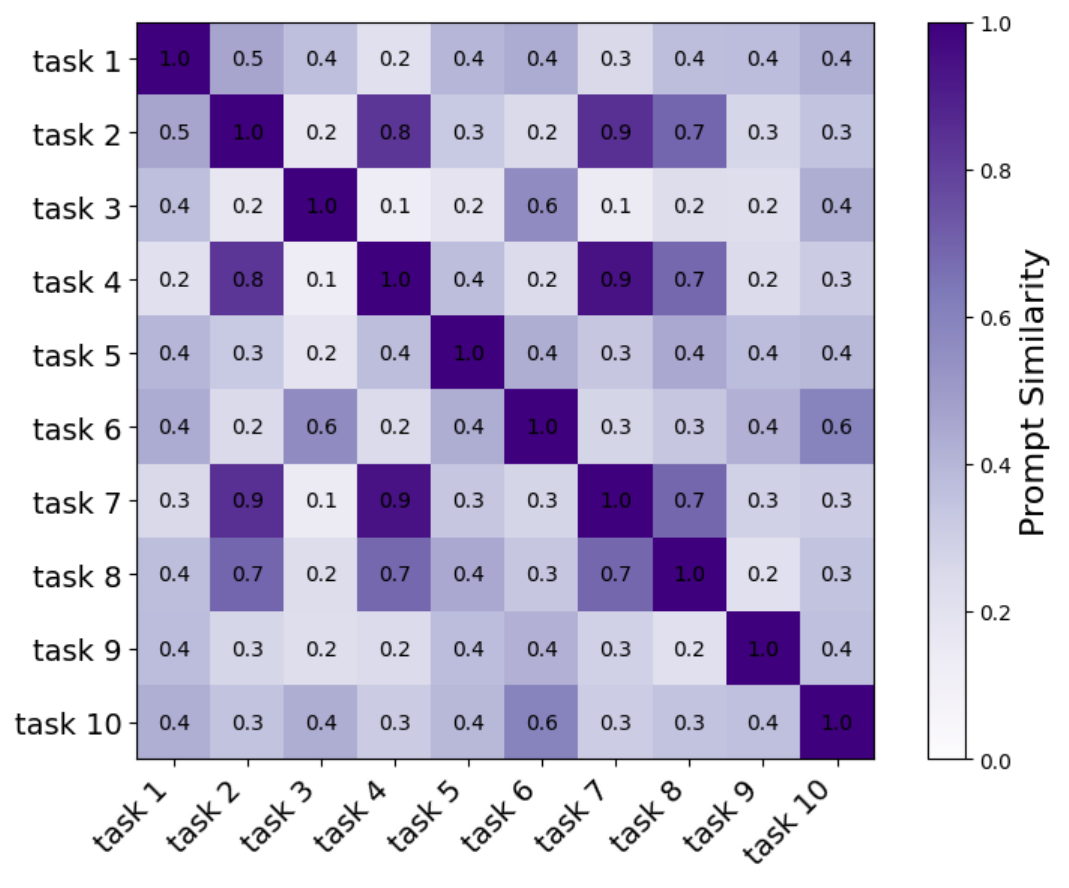}
    }
    \subfigure[Layer 1: $\phi_k^{(1)}$]
    {
        \includegraphics[width=0.3\linewidth]{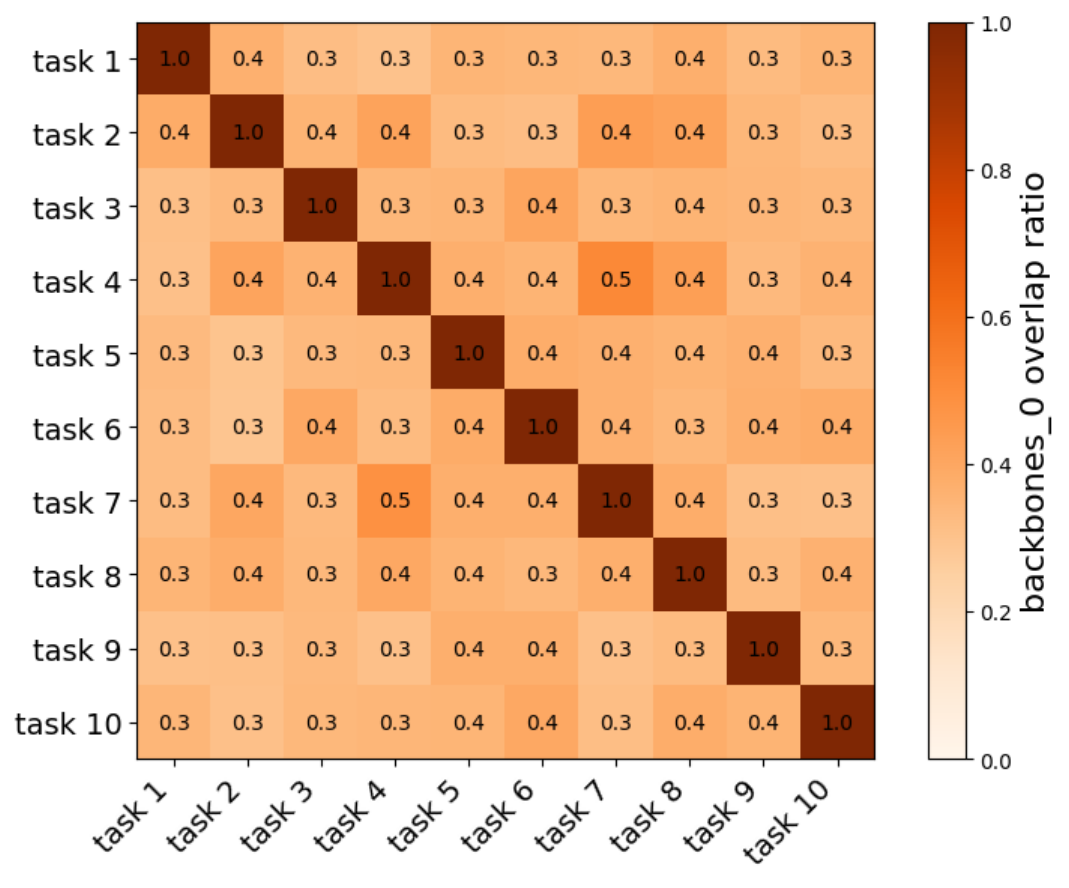}
    }
    \subfigure[Layer 2: $\phi_k^{(2)}$]
    {
        \includegraphics[width=0.3\linewidth]{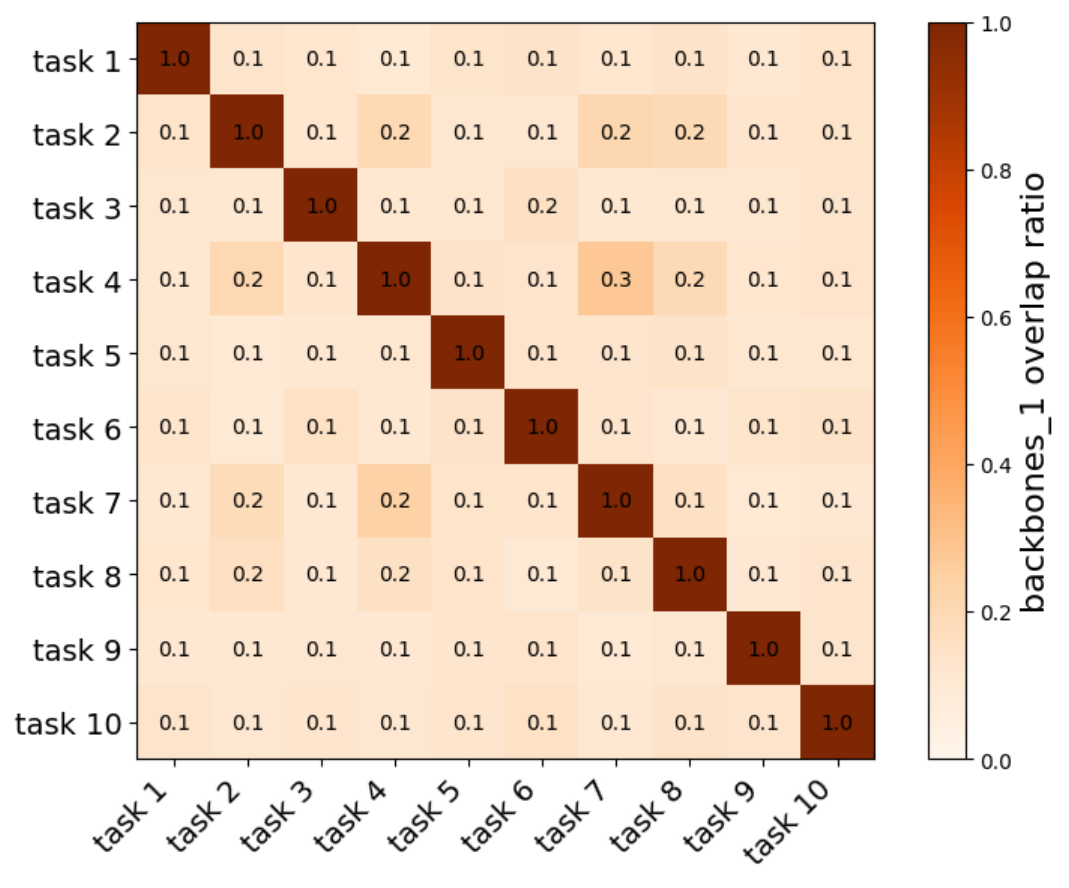}
    }
    \\
    \subfigure[Layer 3: $\phi_k^{(3)}$]
    {
        \includegraphics[width=0.3\linewidth]{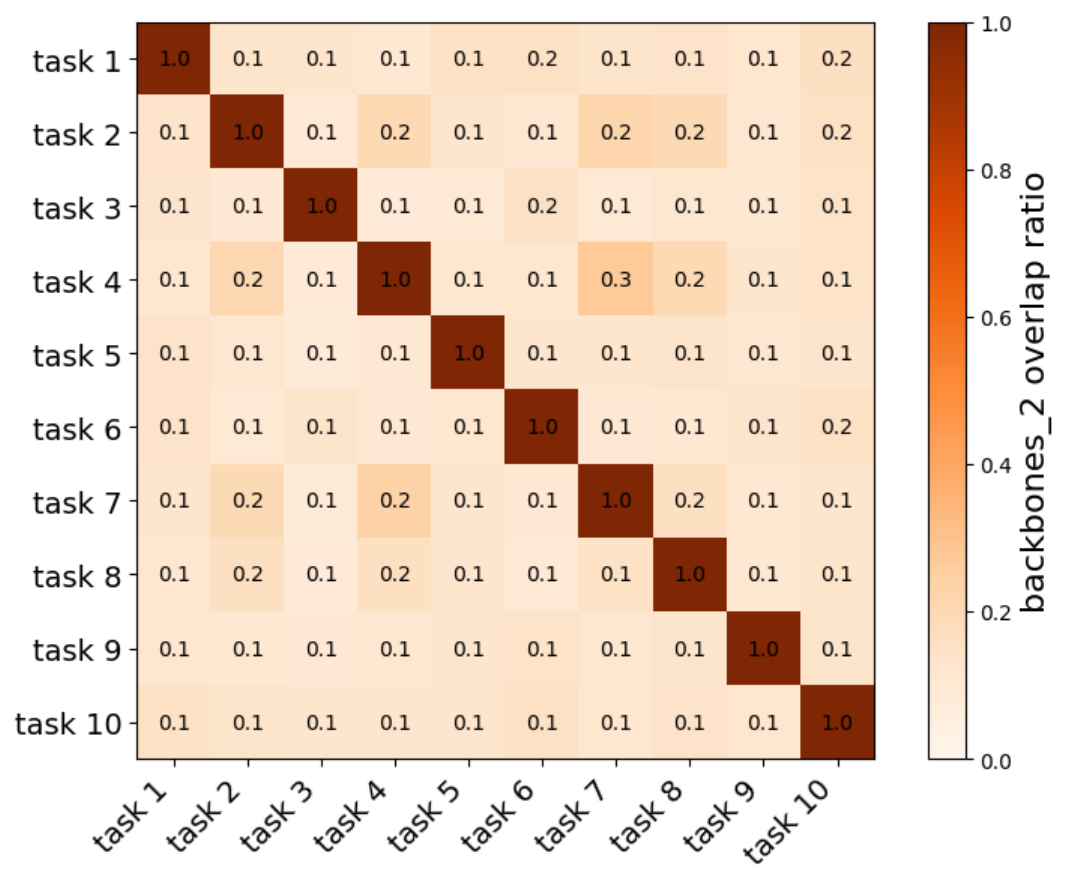}
    }
    \subfigure[Layer 4: $\phi_k^{(4)}$]
    {
        \includegraphics[width=0.3\linewidth]{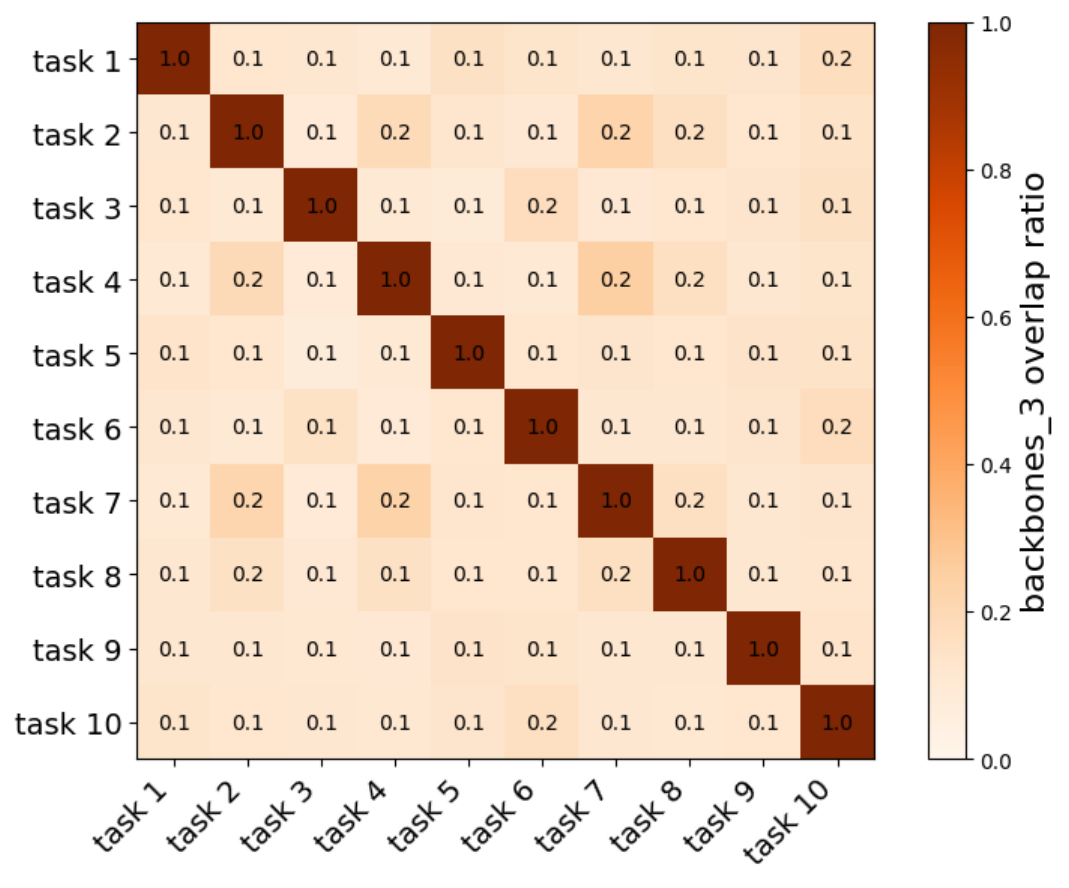}
    }
    \subfigure[Mean\_Layer: $\phi_k^{(\mu)}$]
    {
        \includegraphics[width=0.3\linewidth]{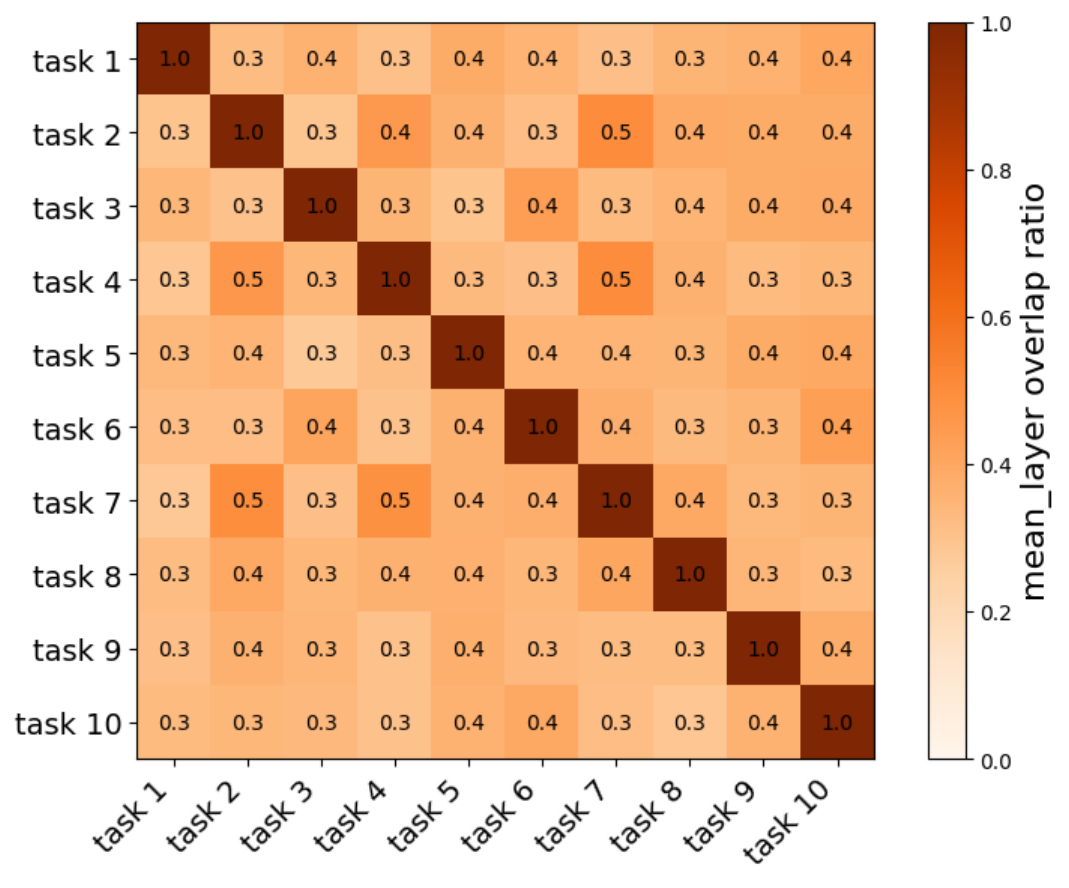}
    }
    \caption{Visualization of \textbf{task description similarity} (a) and \textbf{sub-network mask similarity across layers} (b-f) for SSDE. The strong alignment between the task description heatmap and sub-network allocation masks across layers indicates that SSDE effectively captures task similarities encoded in the descriptions.}
    \label{figure:task_overlap}
\end{figure*}
\begin{figure*}[t!]
    \centering
    \includegraphics[width=1.\textwidth]{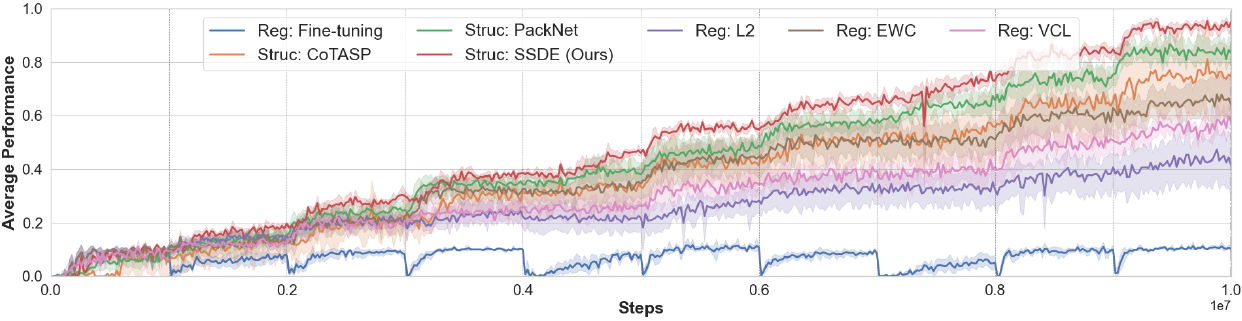} 
    \caption{Learning curves for each method on CW10-v1. By mitigating forgetting through sparse neuron co-allocation and dormant-guided exploration, SSDE significantly outperforms all baseline methods, achieving a task performance of 95\%.}
    \label{fig:leaning_curves}
\end{figure*}

\subsection{Benchmark  Results on Continual World }

\begin{figure*}[t]
    \centering
    \includegraphics[width=1\textwidth]{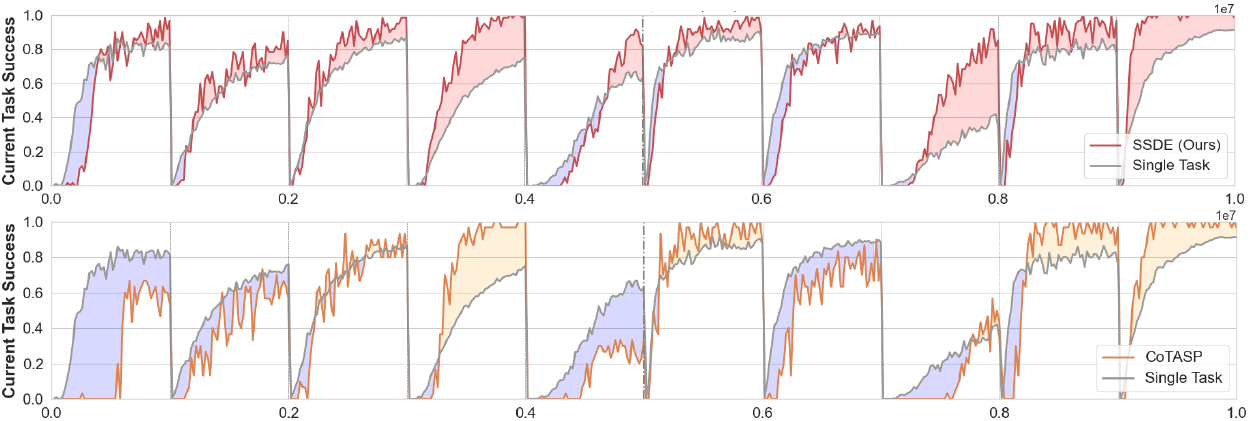} 
    \caption{\textbf{Forward Transfer (\emph{Plasticity}) of SSDE and CoTASP.} Each row compares the learning curve of CW10-v1 tasks against the standard \emph{single-task} SAC baseline provided by the benchmark. \textbf{Blue} regions indicate \textit{negative} transfer, where the single-task baseline learns is faster, while \textbf{Red/Yellow} regions denote \textit{positive} transfer, reflecting high plasticity for continual RL policies. SSDE consistently enables faster and more effective learning than SAC, demonstrating strong \emph{plasticity} and significantly reducing negative transfer compared to its structure-based counterpart, CoTASP.}
    \label{fig:forward_transfer} 
\end{figure*}

We conduct benchmark evaluations on the Continual World 10 Tasks (CW-10) \& 20 Tasks (CW-20) environments, with results presented in Table~\ref{table:main}. 
Overall, {SSDE} demonstrates a superior success ratio on CW10-v1, improving the state-of-the-art record of 86\%, held by a strong rehearsal-based baseline ClonEx-SAC, to 95\%, marking a 9\% increase. It also significantly outperform its strong structure-based counterparts, PackNet and CoTASP.
Figure~\ref{fig:leaning_curves} illustrates the learning curve for SSDE alongside representative baselines. The curve shows a clear advantage for SSDE compare to strong structure-based counterparts like PackNet and CoTASP. Additionally, we demonstrate our method could enhance \emph{plasticity} by showing the forward-transfer effect in Figure~\ref{fig:forward_transfer}, which compares SSDE and CoTASP to a standard single-task policy provided by Continual World. The results highlight that SSDE achieves stable policy learning progress, converging to higher success ratio with positive forward-transfer (\emph{plasticity}).  
\begin{table}[h] 
\caption{Evaluation on  CW10-v2.}
\centering
\label{table:cw10_v2}
\scalebox{1.25}{
\begin{tabular}{lc}
\toprule
\textbf{Method} & \textbf{Average Success $(\uparrow)$} \\
\midrule
CoTASP  & 0.73{\color[HTML]{525252}$\pm$0.13} \\
PackNet & 0.82{\color[HTML]{525252}$\pm$0.04}\\
\textbf{SSDE (Ours) }& \textbf{0.87}{\color[HTML]{525252}$\pm$\textbf{0.03}}\\
\bottomrule
\end{tabular}
}
\end{table}
We also demonstrate SSDE's scalability in handling more tasks through CW20-v1 experiments. {SSDE achieves a comparable performance to ClonEx-SAC, a strong behavior-cloning baseline. It's important to note that CW20 repeats CW10 twice, and ClonEx-SAC would gain access to all expert data and policies for \textbf{all CW20 tasks}, resembling offline RL. Our method treats each task as a \emph{new} task and advances the best score for the structure-based method from 80\% to 87\%. To further illustrate the consistency of SSDE's performance, we evaluated it on CW10-v2. As shown in Table~\ref{table:cw10_v2}, SSDE significantly outperforms its structure-based counterparts CoTASP and PackNet. 
}


To better assess the quality of the sub-networks generated by SSDE, we provide visualization of the similarity heatmaps of sub-network masks allocated by SSDE in Figure~\ref{figure:task_overlap}.
For tasks with similar task embeddings (e.g., Task-2 vs. Task-4 and Task-2 vs. Task-7), we notice strong alignment between the task description heatmap and sub-network allocation masks for each layer. This demonstrates that SSDE effectively captures task similarities encoded in the descriptions.
The strong alignment is crucial for fast adaptation and enhanced \emph{plasticity}, as SSDE can allocate forward-transfer parameters from similar tasks, allowing new tasks to leverage high-quality parameters trained on previous tasks. 
Additional visualizations of the sub-network mask $\phi^{(l)}$ for each layer are provided in Figure~\ref{figure:task_overlap}.

\subsection{Ablation Study}

Table~\ref{table:ablation} presents the results of the ablation study on the CW10-v1 sequence, using average success as the evaluation metric. We evaluate three SSDE variants: 
(1) ``w/o $\beta$'', which disables the Trade-off Coefficient mechanism, keeping all frozen parameters unchanged during training; 
(2) ``w/o Dormant'', which removes the reset parameters mechanism; and 
(3) ``w/o Fine-Grained $\phi, \Psi$'', which relies solely on a fixed dictionary for sub-network allocation. 
Additionally, we introduce an ablation baseline, ``w/o Dormant, w/o $\beta$'', which employs only the co-allocation mask for a fair comparison with the sparse prompting-based allocation from CoTASP. We also include ``SSDE(w/ ReDo)'', comparing our sensitivity-guided dormant score with the original dormant mechanism from ReDo~\citep{Dormant}.

\begin{table}[t]
\caption{Ablation study on {CW10-v1}.}
\centering
\label{table:ablation}
\scalebox{1.25}{
\begin{tabular}{lc}
\toprule
\textbf{Method} & {\textbf{Average Success} ($\uparrow$)} \\
\midrule
{w/o ${\beta}$} & 0.83 $\pm$  0.14\\
{w/o Dormant} & {0.85} $\pm$  0.06\\
{w/o Fine-Grained $\phi,\Psi$} & {0.80} $\pm $ 0.07 \\
{w/o Dormant, w/o ${\beta}$} & {0.81 $\pm$ 0.08} \\
\midrule
{SSDE (w/ ReDo)} & {0.88 $\pm$ 0.02} \\

\textbf{SSDE (ours)} & \textbf{0.95} $\pm$  0.02 \\
\bottomrule
\end{tabular}
}
\end{table}
From Table~\ref{table:ablation}, we observe that the removal of Fine-Grained mask allocation has the most significant impact on the experimental results. Compared to the sparse prompting in CoTASP, our co-allocation improves the performance by more than 10\%. This underscores the critical role of ensuring a dedicated allocation of trainable parameters to incorporate new knowledge, an aspect that has been largely overlooked in previous works.
We also observed that the Trade-off Coefficient $\beta$ contributes to more stable experimental results. This is due to its ability to effectively alleviate the impact of frozen parameters on model performance, leading to more consistent outcomes. 
Additionally, comparing SSDE's sensitivity-guided dormant with ReDo underscores the importance of connecting the sensitivity at observation level to the neuron's activation to address rigidity of sparse policy with particularly constrained trainable capacity.
Overall, the core components of SSDE, each addressing critical aspects of continual RL, work synergistically to form the foundation of the model's success.

\begin{table}[h]
\centering
\caption{Sensitivity analysis of key hyperparameters $\beta$ and $\tau$ on {CW10-v1}.}
\label{table:sensitivity}
\scalebox{1.25}{
\begin{tabular}{lcc}
\toprule
\textbf{Hyperparameter} & \textbf{Value} & \textbf{Average Success ($\uparrow$)} \\
\midrule
\multirow{3}{*}{Dormant Threshold $\tau$} 
& 0.2 & 0.86 $\pm$ 0.01 \\
& 0.6 & \textbf{0.95} $\pm$ \textbf{0.02} \\
& 0.8 & 0.83 $\pm$ 0.06 \\
\midrule
\multirow{4}{*}{Trade-off Coefficient $\beta$} 
& 0.1 & 0.72 $\pm$ 0.01 \\
& 0.3 & \textbf{0.95} $\pm$ \textbf{0.02} \\
& 0.5 & 0.84 $\pm$ 0.05 \\
& 0.8 & 0.83 $\pm$ 0.08 \\
\bottomrule
\end{tabular}
}
\end{table}

We further conducted a sensitivity analysis on the key hyperparameters: dormant threshold $\tau$ and trade-off coefficient $\beta$, as shown in Table~\ref{table:sensitivity}. The results reveal that moderate values of both parameters lead to the most effective learning dynamics. Specifically, $\tau = 0.6$ achieves the highest average success rate of 0.95, highlighting the importance of a well-calibrated reset threshold that restores neuron expressivity while preventing excessive resets that could destabilize learning. Similarly, $\beta = 0.3$ yields the best performance, underscoring the need for a carefully balanced trade-off coefficient that ensures synergy between frozen and trainable parameters. 
Striking the right balance in these hyperparameters is critical for harnessing SSDE’s full potential. An excessively low $\beta$ limits plasticity, restricting the model's ability to adapt, while an overly high $\beta$ introduces instability by over-relying on past policies. Likewise, improper tuning of $\tau$ can either stifle representation learning by failing to reactivate neurons or lead to overly aggressive resets that disrupt knowledge retention.

\section{Conclusion}
We introduce SSDE, a novel structure-based continual RL method designed to balance plasticity and stability in multi-task learning. At its core, SSDE features an efficient co-allocation algorithm that dynamically allocates dedicated capacity for trainable parameters, ensuring task-specific learning while leveraging frozen parameters for efficient forward transfer. This is further balanced by a trade-off parameter for fine-grained inference. To address the expressivity limitations of sparse sub-networks, SSDE incorporates sensitivity-guided dormant neurons, enabling a structural exploration strategy that promotes adaptive representation learning.
These innovations position SSDE as a robust and scalable framework for continual reinforcement learning, with strong potential for tackling multi-task learning, continual adaptation, and continuous control. Looking ahead, further advancements in structural sparsity could enhance sub-network specialization, while integrating differentiable neuron wiring mechanisms offers a promising avenue for improving the expressiveness and adaptability of neural policies. Additionally, exploring SSDE’s potential in diverse application scenarios, such as autonomous systems and lifelong learning settings where tasks continuously evolve, could further enhance its adaptability and impact in real-world reinforcement learning problems.

\bibliography{reference}
\bibliographystyle{IEEEtran}

{
\appendices

\subsection*{\textnormal{APPENDIX}} 
\subsection{Implementation Details} \label{appendix:implementation}
\begin{table*}[b!]

\caption{Detailed hyperparameter configurations for \textbf{SSDE}.}
\begin{center}
\scalebox{1.25}{
\begin{tabular}{lll}
\toprule
\textbf{Hyperparameter} & \textbf{Value}  & \textbf{Range} \\
\midrule
\textbf{SAC} & & \\
\midrule
Actor hidden size & $1024$ &  $\{256, 512, 1024\}$\\
Critic hidden size & $256$ & $\{256, 512, 1024\}$\\
\# of hidden layers for meta policy & $4$ &   $\{2, 3, 4\}$\\
\# of hidden layers for critic $Q_1$ & $4$ &   $\{2, 3, 4\}$\\
\# of hidden layers for critic $Q_2$ & $4$ &   $\{2, 3, 4\}$\\
Activation function & $\text{LeakyReLU}$ &  - \\
Batch size & $256$ &  \{64, 128, 256\}  \\
Discount factor & 0.99 & -\\
Target entropy & $-2.0$ & -\\
Target interpolation &$5\times10^{-3}$ & -\\
Replay buffer size & 1e6 & \{2e5,5e5,1e6\}\\
Exploratory steps & 1e4 & -\\
Optimizer & Adam & -\\
Learning rate & $3\times10^{-4}$ & -\\
\midrule
\textbf{Continual Learning} & & \\
\midrule
Training steps for each task & 1e6 & \\
Evaluation steps for each task & & \\
\midrule
\textbf{SSDE} & & \\
\midrule
Sparsity ratio $\lambda_{\Gamma}, \lambda_{\Lambda}$ & $10^{-3}$ & $\{10^{-2}, 10^{-3}, 10^{-4}, 10^{-5}\}$  \\
Trade-off parameter $\beta$ & $0.3$ & $\{0.1, 0.2, 0.3, 0.4, 0.5, 0.6, 0.7, 0.8, 1\}$ \\
Dormant threshold $\tau$ & $0.6$  & \{0.2, 0.4, 0.6, 0.8\} \\
Dormant reset interval & $8e4$  & \{1e4, 2e4, 5e4, 8e4, 1e5, 2e5\}\\
\bottomrule
\end{tabular}
}
\label{table:hyperparameters}
\end{center}
\end{table*} 

SSDE is developed on top of the Jax Implementation of SAC from JaxRL. The actor policy and critic networks are parameterized as standard MLPs. Notably, SSDE introduces no additional trainable parameters compared to SAC. All calibration masks are determined beforehand through the co-allocation strategy, ensuring that during training, the masks remain fixed, allowing for efficient learning with pre-allocated sub-networks. As a result, SSDE achieves highly computationally efficient sub-network allocation. SSDE also does not employ task-specific policy heads. Additionally, we do not store any data from previous tasks or perform rehearsal on past experiences, differentiating it from rehearsal-based approaches like ClonEx-SAC. 
For the evaluation on computational efficiency presented in Table~\ref{tab:mask_allocation_time}, we use a GPU server with L40 GPU, and $120$ cores ``AMD EPYC 9554P 64-Core Processor'' CPU. 

\subsection{A Case Study on Sensitivity-Guided Dormant Neurons}
 \label{apendix:case_study_T5}

\begin{figure*}[t]
  \begin{center}
    \centering
    \subfigure[Successful Policy]{
        \includegraphics[width=0.31\linewidth]{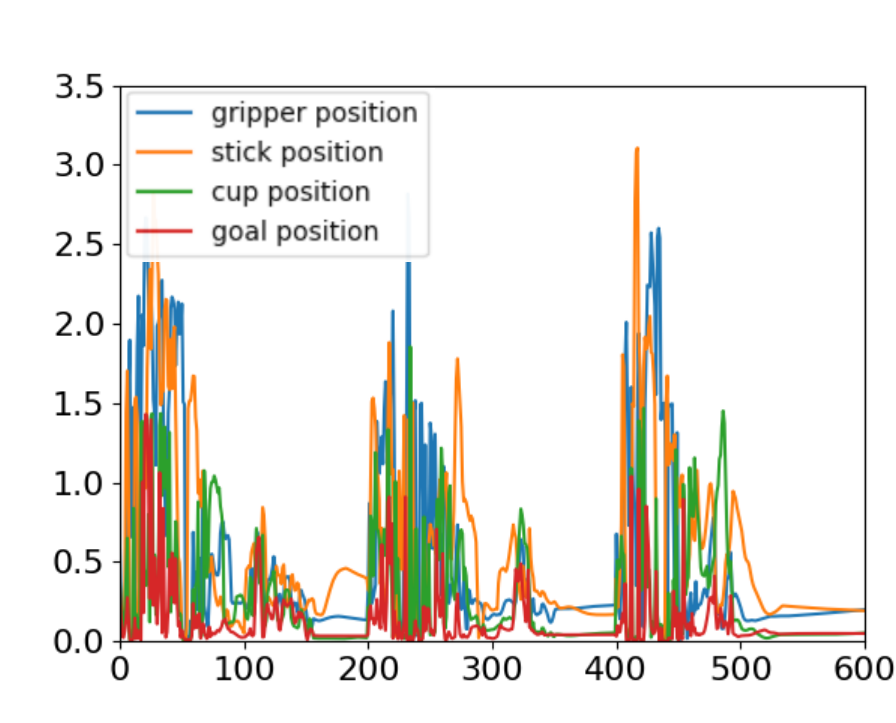}
    }
    \subfigure[Failed Policy]{
        \includegraphics[width=0.31\linewidth]{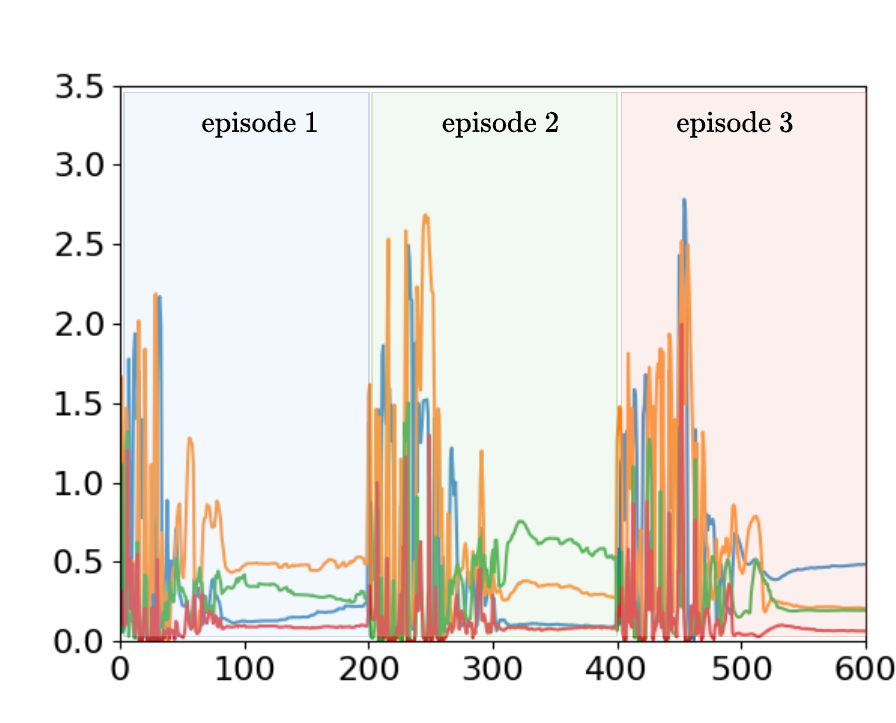}
    }
    \subfigure[Learning curve on Task-5]{
        \includegraphics[width=0.31\linewidth]{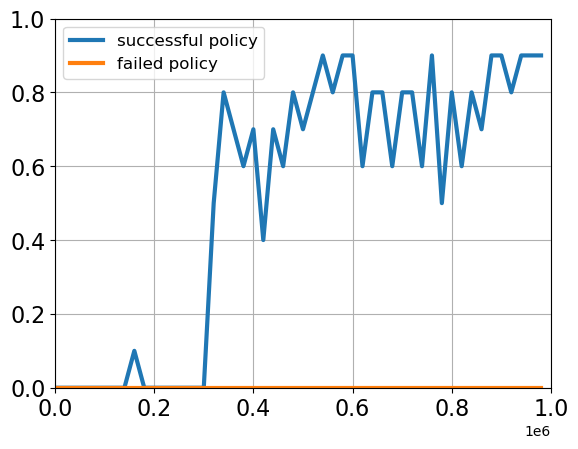}
    }
  \end{center}
  \caption{
  {Evaluation on \textbf{input sensitivity} for a successful policy and failed policy, on Task-5. The failed policy remains insensitive to change on \textbf{goal position}, compared to other features, and thereby fail in pushing the cup to the goal.}
  }
  \label{figure:input_sensitivity}
\end{figure*}
\begin{figure}[h!]
  \begin{center}
    \vskip -0.1in
    \centering
    \subfigure[successful Policy]{
        \includegraphics[width=\linewidth]{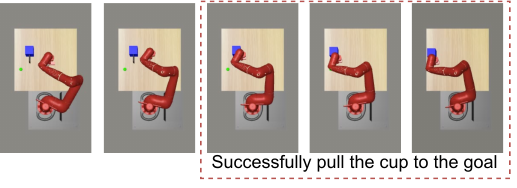}
    }
    \subfigure[Failed Policy]{
        \includegraphics[width=\linewidth]{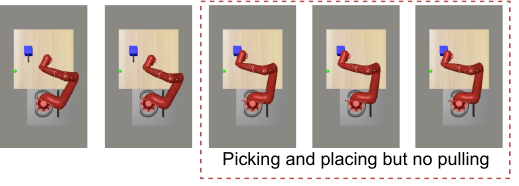}
    }
  \end{center}
  \caption{Demonstration of the rigidity in the sparse continual RL policy. The failed policy (\textbf{bottom}) stuck in sub-optimal solutions: the agent can \emph{pick up stick}, \emph{place stick to cup}, but cannot \emph{pull cup to goal location}.\textbf{ The failed policy is highly insensitive to change on goal location from the state inputs.}}
  \label{figure:task_5_trajectory}
\end{figure}

Sparse sub-policies in structure-based continual RL methods often encounter the challenge of \textbf{rigidity}, where limited trainable parameters progressively lose sensitivity to input variations. This rigidity significantly hinders the policy's ability to adapt to complex tasks requiring precise, multi-step execution. Traditional approaches like \textbf{ReDo}~\citep{Dormant}, which assess neuron responsiveness based solely on activation scales, inadequately capture this issue, as they overlook the critical role of neuron sensitivity to input variations. In this section, we use \textbf{Task-5 Stick-Pull} as a case study to statistically measure and showcase the rigidity of sparse sub-policies in terms of input sensitivity. Our observations reveal that policies with restrictive expressivity, constrained by sparse allocation, fail to respond effectively to critical state inputs. This limitation motivates the development of our \textbf{sensitivity-guided dormant scores}, which bridge neuron activation with input sensitivity. By addressing the expressivity challenges inherent in sparse sub-policies, SSDE enhances their responsiveness and adaptability, enabling them to handle such demanding tasks more effectively.

\subsubsection{Task Description} \textbf{Task-5 Stick-Pull} (\textit{Grasp a stick and pull a cup with the stick}) is a relatively complex manipulation task from CW10.  It involves multiple sub-skills, which need to be executed in sequence: (1) maneuvering the arm to the stick, (2) picking it up, (3) placing the stick into a hole, and (4) finally pulling the cup to a designated target position. Many continual RL policies struggle with this task, often resulting in sub-optimal policies and failing to achieve desired goal-reaching performance. Unless the agent successfully manages all sub-skills, the policy will score a 0\% success rate. While training a large network with SAC can easily yield a 100\% success rate, structure-based continual learning methods, such as CoTASP, often perform poorly, frequently resulting in failed policy with a 0\% success rate on this task.

\subsubsection{Analysis}  Although the environment provides dense rewards, the failed agent is only able to learn partial skills. As shown in Figure~\ref{figure:task_5_trajectory}, the agent successfully picks up and places the stick but fails to pull it toward the goal. This illustrates a common scenario where a policy becomes stuck in a sub-optimal solution due to insufficient exploration. Enhancing the exploration capabilities of the sparse sub-network remains a critical challenge to address.
Additionally, we pose the following {assumption}: 

{\textbf{Q}: \emph{Could the failure for sparse sub-networks to learn complex manipulation skills be attributed to the \textbf{insensitivity of the sub-network parameters to changes in key input features}?}

\subsubsection{Empirical Results} \textbf{We empirically evaluated the \emph{sensitivity} for sub-network policy parameters w.r.t changes in input. }We grouped the input into four major categories (\textit{gripper position, stick position, cup position, and goal position}), and applied $\Delta x$ 
to each group, where $\Delta x$  is a static perturbation noise to the input. The change in the network's response to the perturbed state is denoted as $\va'$, and the difference $|\va' - \va|$ is recorded. We present the sensitivity of a successful policy and a failed policy, from Figure~\ref{figure:input_sensitivity} (a) and (b), respectively. The results show that, while the failed policy exhibits high sensitivity to \emph{cup position} features and \emph{stick position},  its sensitivity to \emph{goal} is significantly lower. In contrast, the successful policy demonstrates more balanced sensitivity across all feature groups, without overlooking certain inputs like goal position. 
\textbf{This motivates us to propose a sensitivity-guided exploration strategy.} We define a novel concept of \textbf{\emph{sensitivity}-guided dormant score,} using it to actively identify insensitive parameters in response to input perturbations.

\end{document}

%% file: math_commands.tex
\usepackage{amsmath,amsfonts,bm}

\def\eqref#1{equation~\ref{#1}}

\def\1{\bm{1}}

\def\va{{\bm{a}}}

\def\vg{{\bm{g}}}

\def\vs{{\bm{s}}}

\def\vy{{\bm{y}}}

\def\mD{{\bm{D}}}

\def\mW{{\bm{W}}}

\DeclareMathAlphabet{\mathsfit}{\encodingdefault}{\sfdefault}{m}{sl}
\SetMathAlphabet{\mathsfit}{bold}{\encodingdefault}{\sfdefault}{bx}{n}

\DeclareMathOperator*{\argmin}{arg\,min}